\documentclass[runningheads]{llncs}

\usepackage[T1]{fontenc}
\usepackage{graphicx,xcolor}
\usepackage{amsfonts}
\usepackage{amsmath}
\usepackage{amssymb}
\usepackage{algorithm}
\usepackage{algpseudocode}
\usepackage{hyperref}
\usepackage{booktabs}
\usepackage{multirow}
\usepackage{enumitem}
\usepackage{makecell}
\usepackage{multirow}

\algtext*{EndWhile}
\algtext*{EndFor}
\algtext*{EndIf}

\begin{document}

\allowdisplaybreaks

\title{Solving 0-1 Integer Programs with Unknown Knapsack Constraints Using Membership Oracles}

\titlerunning{Solving 0-1 Integer Programs with Unknown Knapsack Constraints}

\author{Rosario Messana, Rui Chen, Andrea Lodi, Alberto Ceselli}

\authorrunning{Messana et al.}


\maketitle

\begin{abstract}
We consider solving a combinatorial optimization problem with unknown knapsack constraints using a membership oracle for each unknown constraint such that, given a solution, the oracle determines whether the constraint is satisfied or not with absolute certainty. The goal of the decision maker is to find the best possible solution subject to a budget on the number of oracle calls. Inspired by active learning for binary classification based on Support Vector Machines (SVMs), we devise a framework to solve the problem by learning and exploiting surrogate linear constraints. The framework includes training linear separators on the labeled points and selecting new points to be labeled, which is achieved by applying a sampling strategy and solving a 0-1 integer linear program. Following the active learning literature, a natural choice would be SVM as a linear classifier and the information-based sampling strategy known as simple margin, for each unknown constraint. We improve on both sides: we propose an alternative sampling strategy based on mixed-integer quadratic programming and a linear separation method inspired by an algorithm for convex optimization in the oracle model. We conduct experiments on classical problems and variants inspired by realistic applications to show how different linear separation methods and sampling strategies influence the quality of the results in terms of several metrics including objective value, dual bound and running time.

\keywords{Combinatorial optimization \and Oracle-based optimization \and Active learning.}
\end{abstract}

\section{Introduction}
\label{sec:Introduction}

Mathematical programming is a powerful way of modeling and solving optimization problems. It has been proven to be effective for solving optimization problems with hard (e.g., combinatorial) constraints, in particular with help from off-the-shelf solvers (e.g., Gurobi \cite{gurobi}). The formalization of an optimization problem into a correct mathematical programming model is a process that requires expertise and knowledge about the specific application field. Lack of such resources or the intrinsic nature of the problem can make it difficult or impossible to obtain a fully specified mathematical model by applying the traditional modeling approach that relies extensively on human intervention. Data-driven techniques can help reduce this gap between applicability of mathematical programming and modeling capability, either by enabling solution techniques that do not require an explicit representation of objectives and constraints (e.g., \cite{dadush2023simple,lee2015faster}) or by using machine learning methods to automate the formulation of mathematical programming models (e.g., \cite{fajemisin2024optimization,lombardi2018boosting}).

In this paper, we take a step inside the development of specific techniques for combinatorial optimization in an interactive setting. We are interested in solving combinatorial optimization problems that are only partially formalized while learning a representation of the non-described portion. In order to limit the computational complexity of the optimization task, we assume $m$ unknown knapsack constraints, i.e., constraints in less or equal form whose parameters are non-negative.
We rely on a source of feedback consisting in receiving yes/no feasibility responses on single solutions with respect to each unknown constraint and we assume the feedback to be correct with absolute certainty. More formally, we want to solve
\begin{subequations}
\label{model:01LinearProgramWithKnapsackConstraints}
\begin{align}
    \max~ & z(\boldsymbol{x}) := \sum_{i=1}^m \sum_{j=1}^n v_{ij} \cdot x_{ij} \label{obj:01LinearProgramWithKnapsackConstraints} \\
    \text{s.t.}~ & \sum_{j=1}^n \bar{w}_{ij} \cdot x_{ij} \leq 1, & i \in I, \label{constr:KnapsackConstraints} \\
    & \boldsymbol{x} \in X \subseteq \{0, 1\}^{m \times n}. \label{constr:Combinatorial}
\end{align}
\end{subequations}
For every $i \in I$, the weights $\bar{\boldsymbol{w}}_i$ of the corresponding constraint in \eqref{constr:KnapsackConstraints} are unknown. We assume them to belong to the $n$-dimensional unit cube, i.e. $\bar{\boldsymbol{w}}_i \in W_i := [0, 1]^n$, but we also allow to restrict $W_i$ by cutting the cube with a finite number of linear inequalities in order to inject additional prior knowledge on the weights. The set $X$ encodes the known combinatorial structure of the problem in addition to the integrality constraints.
Let us define $W:=\prod_{i\in I}W_i$ and, for each $i \in I$, $X_i:=\text{proj}_{\boldsymbol{x}_i}(X)$.
For each $i \in I$, the unknown constraint $i$ in \eqref{constr:KnapsackConstraints} is characterized by an individual \emph{membership oracle} that, given $\boldsymbol{\chi} \in X_i$, can determine whether $\boldsymbol{\chi}$ satisfies the constraint or not. We refer to any $\boldsymbol{\chi} \in X_i$ as a \emph{sub-solution} of (\ref{model:01LinearProgramWithKnapsackConstraints}) with respect to the unknown constraint $i$. Non-empty initial sets $S^+ \subseteq X$ and $S^- \subseteq X$ are given, containing feasible and infeasible solutions to \eqref{constr:KnapsackConstraints}, respectively.
Finally, we are given a maximum number $N$ of oracle calls that applies to each oracle separately. We highlight that, even if the unknown constraints \eqref{constr:KnapsackConstraints} have pairwise disjoint sets of variables, Problem \eqref{model:01LinearProgramWithKnapsackConstraints} is general enough to account for constraints with overlapping sets of variables, as one can add to $X$ constraints of the form $x_{ij} = x_{i^{\prime}j^{\prime}}$ to account for the fact that $x_{ij}$ and $x_{i^{\prime}j^{\prime}}$ coincide. \\

From a theoretical standpoint, the complexity of optimizing problem \eqref{model:01LinearProgramWithKnapsackConstraints} in terms of number of oracle calls, even with a single unknown constraint, is exponential in the dimension $n$. The reader may refer to Theorem \ref{thm:HardnessResult} for a complexity result in the worst case scenario. Our goal is to find a practical algorithm that computes the best solution to the problem, say $\hat{\boldsymbol{x}}$, within the given number $N$ of oracle calls. Inspired by active learning methods for binary classification based on Support Vector Machines (SVMs) \cite{kremer2014active}, we define an iterative solution paradigm prescribing some key steps. The first one is a \emph{linear separation} step that consists in training a linear classifier for each unknown constraint to separate the known sub-solutions that satisfy the constraint from those violating it, and to find the equation of the separating hyperplane $h_i$ defined by $\hat{\boldsymbol{w}}_i \cdot \boldsymbol{x}_i = 1$, with $\hat{\boldsymbol{w}}_i \in W_i$. In a second phase, we apply a suitable \emph{sampling strategy} to obtain a previously unseen sub-solution for each unknown constraint. Each sub-solution is submitted to the respective membership oracle to know if it satisfies the corresponding constraint. Then, we solve the 0-1 integer programming (IP) model using for each $i \in I$ the surrogate knapsack constraint $\hat{\boldsymbol{w}}_i \cdot \boldsymbol{x}_i \leq 1$ in place of the unknown one to select a solution. The optimal solution of the surrogate 0-1 IP model is split into sub-solutions, one for each unknown constraint, and each sub-solution is submitted to the respective oracle. Our solution framework maintains both an upper and a lower bound on the value of an optimal solution, and terminates whenever their gap is under a certain  or the maximum number of oracle calls is reached. \\

After introducing the general framework, we discuss specific solution approaches within the framework. First, according to an established approach in active learning for binary classification, we consider using a SVM for linear classification together with a related sampling strategy that is function of the current separating hyperplane (a so-called \emph{information-based} strategy). In particular, we employ the strategy known as \emph{simple margin} \cite{tong2001active}, which is well suited for our combinatorial setting. Then, we propose an alternative sampling strategy called \emph{closest cutting plane} that exploits the discrete nature of the problem and the structure of the so-called \emph{version space} for each sub-problem. 
Finally, we propose an alternative linear separation method obtained as a variant of an algorithm for convex optimization with a \emph{separation oracle} \cite{dadush2023simple}. \footnote{By separation for a separation oracle, we mean finding a hyperplane that separates a point from a convex set, which differs from the notion of separation in linear separation of labeled points.} 
\\

In the experiments, we focus on a set of specific combinatorial problems. In order to analyze the performance of our methods in a basic scenario, we test them on two problems with a single unknown constraint, namely the classic 0-1 knapsack problem and a college study plan problem based on real data. Then, we test on instances of the generalized assignment problem \cite{ross1975branch}, where we assume the capacity constraints to be unknown. For each test problem, we show the relative performance of the different combinations of pairs (separation method, sampling strategy) in terms of number of oracle calls, primal-dual gap and running time. \\

The remainder of the paper is organized as follows. In Section \ref{sec:BackgroundAndLiteratureReview}, we give some background and we review the related literature. In Section \ref{sec:InteractiveSamplingEnhancedOptimization}, we give the details of our solution approaches. In Section \ref{sec:TestProblems}, we present our test problems. In Section \ref{sec:DataCollectionAndGeneration}, we describe how we obtain the test instances for our experiments. In Section \ref{sec:ExperimentalSettingAndResults}, we present our experimental setting and empirical results. Finally, we draw our conclusions in Section \ref{sec:Conclusions}.

\section{Background and Literature Review}
\label{sec:BackgroundAndLiteratureReview}

Our work intersects with existing literature on optimization in partial information settings and machine learning, while also introducing key distinctions outlined in the following.

\subsection{Optimization Under Partial Information}

Specific areas in optimization have traditionally been devoted to solving problems under uncertainty. For example, stochastic programming \cite{fouskakis2002stochastic,powell2019unified} faces problems where probabilistic distributions model unknown parameters, and robust optimization \cite{ben2009robust} revolves around finding solutions that remain feasible and near-optimal even if the parameters change in given ranges. Sometimes the focus is on learning or completing an optimization model by using solution examples, without necessarily solving the underlying problem. It is the case, for instance, of model acquisition and constraint acquisition, which have seen advancements in both static \cite{kudla2018one,kumar2019acquiring,pawlak2017automatic} and interactive settings \cite{bessiere2023learning,mechqrane2024using,zheng2017active}. Another notable example is inverse optimization \cite{chan2025inverse}, in which the goal is determining the parameters of an optimization problem that make a given solution optimal.
To the best of our knowledge, no existing works target 0-1 integer programming with unknown constraints in an interactive setting. However, our work can be situated within oracle-based optimization. This research area has received substantial attention over the years for its role in interactive applications where an explicit description of the feasible region is not available or is difficult to formalize. A well-studied setting consists in minimizing a convex function $f: \mathbb{R}^d \rightarrow \mathbb{R}$ over a convex set $K$ equipped with a separation oracle \cite{bertsimas2004solving,cheney1959newton,dadush2023simple,goffin1993computation,gomory2010outline,kelley1960cutting,lee2015faster,nesterov1995cutting,vaidya1996new}, which is proven to be solvable with a pseudo-linear number of oracle calls with respect to the dimension $d$. Recent advances include convex optimization with a membership oracle for $K$ and an evaluation oracle for $f$ \cite{lee2018efficient}. In the case of convex sets, separation and membership oracles are equivalent from a complexity standpoint. Indeed, according to the classical result in \cite{grotschel2012geometric}, there exists a polynomial time equivalence between optimization, separation and membership for convex bodies. Since in this paper we are interested in the combinatorial optimization scenario, convenient theoretical results like those described above are not available due to the non-convexity of the feasible region. Actually, as we show in Theorem \ref{thm:HardnessResult}, any algorithm requires $\Omega(n^{1/\epsilon})$ calls to the membership oracle in expectation to find a $(1-\epsilon)$-approximate solution of Problem \eqref{model:01LinearProgramWithKnapsackConstraints} with one unknown knapsack constraint in the worst case. The proof can be found in Appendix \ref{sec:Proofs}.
\begin{theorem}
\label{thm:HardnessResult}
    For any deterministic algorithm for Problem \eqref{model:01LinearProgramWithKnapsackConstraints} with a single unknown constraint, for any $\epsilon\in(0,1)$, there exists an instance of the problem that takes $\Omega(n^{1/\epsilon})$ oracle calls to label a feasible solution whose objective value is within a multiplicative factor of $(1-\epsilon)$ with respect to the optimal solution.
\end{theorem}

\subsection{Active learning for binary classification}
The solution approaches proposed in this paper to solve the general problem introduced in Section \ref{sec:Introduction} are based on learning surrogate linear constraints or, in other terms, an approximation of the halfspace containing the valid sub-solutions according to each unknown constraint. Halfspace learning is a research area with a long tradition that traces back to the perceptron algorithm \cite{rosenblatt1958perceptron}. It has received extensive interest over the years (see, e.g., \cite{awasthi2015efficient,birnbaum2012learning,blum1998polynomial,daniely2015ptas,daniely2016complexity,guruswami2009hardness,shalev2011learning}), especially regarding popular algorithms like, e.g., Support Vector Machines \cite{boser1992training,mammone2009support,vapnik1995support}. A significant part of the work in halfspace learning has been done in the context of active learning \cite{fu2013survey,hino2020active,kumar2020active,settles2009active,tharwat2023survey}. Compared to traditional supervised learning techniques that can usually require massive amounts of labeled data to produce models with desired generalization capabilities, active learning algorithms employ relatively small initial labeled datasets and are based on querying an oracle on new unlabeled data in order to make the learning process less data consuming \cite{tharwat2023survey}. A particular sub-category of active learning methods is represented by algorithms whose distinctive characteristic is to rely on a fixed dataset of unlabeled data and to implement a policy, called \emph{sampling strategy}, to decide in a sequential fashion what unlabeled available points submit to the oracle for labeling. An established active learning method for binary classification consists in alternating the training of a binary classifier on the dataset of already labeled points and a sampling step. The classification task is traditionally accomplished using SVM \cite{ho2011active,schohn2000less,tong2001support,xu2003representative}, while several related sampling strategies have been proposed in the literature (see, e.g., \cite{kremer2014active,tharwat2023survey}). Existing strategies fall in one of three categories: \emph{information-based} strategies, sampling the most informative points, for example selecting those whose label is the most uncertain; \emph{representation-based} strategies, trying to diversify the areas where the points are sampled; and finally, strategies that combine the two previous approaches.

In the presence of separable data, the set of all linear classifiers that correctly separate the current observations is called \emph{version space} \cite{mitchell1982generalization,tharwat2023survey}. Training a classifier on the current labeled points means finding a point inside the current version space. Sampling an unlabeled point and acquiring its label from the oracle corresponds to possibly reducing the version space in size, since every new labeled point either restricts the set of classifiers that can separate the current labeled data or leaves it unchanged. The idea behind alternating between training a linear classifier and  sampling is that a convenient combination of the two steps can shrink the version space significantly, thus facilitating the convergence towards better classifiers for the binary classification problem. As we show in Section \ref{sec:InteractiveSamplingEnhancedOptimization}, our framework relies on multiple classification actions and therefore on the existence of different version spaces that evolve during execution. Moreover, classification is not the goal in itself, but serves as a step in the overall optimization process.

\section{Interactive Sampling-Enhanced Optimization}
\label{sec:InteractiveSamplingEnhancedOptimization}

We here formalize how a combination of separation, sampling and optimization can be applied in our interactive setting to solve Problem \eqref{model:01LinearProgramWithKnapsackConstraints}, resulting in a general framework for combinatorial optimization with unknown knapsack constraints that exploits membership oracles. Since we are interested in obtaining information about the quality of the solutions returned by the algorithm, we maintain an upper bound on the value of the best known feasible solution over the iteration process and we require the relative gap between that bound and the best known objective to be under a given threshold for the algorithm to terminate, unless the maximum number of calls is reached for any of the membership oracles. The framework, that we call \emph{Interactive Sampling-Enhanced Optimization} (ISEO), is presented in Algorithm \ref{alg:ActiveLearnAndOptimize}. The input parameters are the maximum number of calls for each oracle, the initial sets of feasible and infeasible solutions to Problem \eqref{model:01LinearProgramWithKnapsackConstraints}, the domains $W_i$ that constrain the choice of the surrogate weights for every $i \in I$, the objective function \eqref{obj:01LinearProgramWithKnapsackConstraints} and the gap threshold involved in the stopping criterion. For each time step $t$, the pair of sets $S^+(t)$, $S^-(t) \subseteq X$ contain the labeled feasible and infeasible solutions to the target problem, respectively. Moreover, for each $i \in I$, the sets $S^+_i(t)$, $S^-_i(t) \subseteq X_i$ contain the labeled sub-solutions relative to the $i$-th unknown constraint.
The pseudo-code employs the following notation:
\begin{enumerate}
    \item $\mathcal{S}(W_i, S^+_i(t-1), S^-_i(t-1))$ denotes the linear separation algorithm of choice, that outputs $\hat{\boldsymbol{w}}^{t}_i \in W_i$ such that $\hat{\boldsymbol{w}}^{t}_i \cdot \boldsymbol{s}^+_i \leq 1$ for every $\boldsymbol{s}^+_i \in S^+_i(t - 1)$ and $\hat{\boldsymbol{w}}^{t}_i \cdot \boldsymbol{s}^-_i \geq 1$ for every $\boldsymbol{s}^-_i \in S^-_i(t - 1)$;
    \item $\mathcal{T}\left(X_i, S^+_i(t-1), S^-_i(t-1), \hat{\boldsymbol{w}}^{t}_i\right)$ corresponds to the chosen sampling strategy, that returns a point $\boldsymbol{u}^{t}_i$ from $X_i$ such that $\boldsymbol{u}^{t}_i \nleq \boldsymbol{s}^+_i$ for every $\boldsymbol{s}^+_i \in S^+_i(t-1)$ and $\boldsymbol{u}^{t}_i \ngeq \boldsymbol{s}^-_i$ for every $\boldsymbol{s}^-_i \in S^-_i(t-1)$;
    \item $\mathcal{O}\left(X, \hat{\boldsymbol{w}}^t, S^+(t - 1), \left\{S^-_i(t - 1)\right\}_{i=1}^m, z\right)$ is the optimization step consisting in solving the following model:
    \begin{subequations}
    \label{model:OptimizationStep}
    \begin{align}
        \max~ & z(\boldsymbol{x}) \\
        \text{s.t.}~ & \hat{\boldsymbol{w}}^t_i \cdot \boldsymbol{x}_i \leq 1, & i \in I, \\
        & \sum_{i=1}^m \sum_{j=1}^n x_{ij} \cdot (1 - s^+_{ij}) \geq 1, & \boldsymbol{s}^+ \in S^+(t - 1), \\
        & \sum_{j=1}^n x_{ij} \cdot \sigma_j^- \leq \sum_{j=1}^n \sigma_j^- - 1 & \boldsymbol{\sigma}^- \in S^-_i(t - 1),\ i \in I, \\
        & \boldsymbol{x} \in X. \label{constr:Combinatorial}
    \end{align}
    \end{subequations}
    \item $\Omega_i(\boldsymbol{\mu})$ indicates a query on point $\boldsymbol{\mu} \in X_i$ that yields its label $\ell_{\boldsymbol{\mu}} \in \{1, -1\}$, where $\ell_{\boldsymbol{\mu}} = 1$ if and only if $\boldsymbol{\mu}$ satisfies the $i$-th unknown constraint;
    \item $\mathcal{B}\left(X, \left\{W_i\right\}_{i=1}^m, S^+(t), S^-(t), \left\{S^-_i(t)\right\}_{i=1}^m, z\right)$ stands for finding an upper bound $U\!B$ to the value of the current best solution in $S^+(t)$.
\end{enumerate}

\begin{algorithm}
\caption{Interactive Sampling-Enhanced Optimization}
\label{alg:ActiveLearnAndOptimize}
\begin{algorithmic}[1]
\State \textbf{Input:} maximum number of queries per oracle $N$, initial non-empty sets $S^+(0)$ and $S^-(0)$ of labeled points, surrogate weight domains $W_i$ for every $i \in I$, objective function $z(\boldsymbol{x})$, gap threshold $thr$
\State \textbf{Initialize} $t\leftarrow 0$, $flag \gets$ \textbf{true}, $U\!B \gets +\infty$, $L\!B \gets \max\left\{z(\boldsymbol{x})\ |\ \boldsymbol{x} \in S^+(0)\right\}$; for every $i \in I$: $S^+_i(0) \gets \{\boldsymbol{s}_i\ |\ \boldsymbol{s} \in S^+(0)\} \cup \{\boldsymbol{s}_i\ |\ \boldsymbol{s} \in S^-(0) \text{ and } \Omega_i(\boldsymbol{s}_i) = 1\}$, $S^-_i(0) \gets \{\boldsymbol{s}_i\ |\ \boldsymbol{s} \in S^-(0) \text{ and } \Omega_i(\boldsymbol{s}_i) = -1\}$, $n_i \gets |S^-(0)|$ 
\While {$(U\!B - L\!B) / L\!B > thr$ \textbf{and} $flag$ \textbf{is true}} \Comment{Stopping criterion} \label{line:stopping} 
    \State $t \gets t + 1$
    \For {$i \in I$}
        \State $\hat{\boldsymbol{w}}^{t}_i \gets \mathcal{S}(W_i, S^+_i(t-1), S^-_i(t-1))$ \Comment{Linear separation} \label{line:separation}
        \State $P^+_i(t) \gets \emptyset$; $P^-_i(t) \gets \emptyset$ 
    \EndFor
    \For {$\theta \in \{\text{sampling}, \text{optimization}\}$}
        \If {$n_i \geq N$ for any $i \in I$}
            \State $flag \gets$ \textbf{false}
            \State \textbf{break}
        \EndIf
        \If {$\theta = \text{sampling}$}
            \For {$i \in I$}
                \State $\boldsymbol{u}^{t}_i \gets \mathcal{T}\left(X_i, S^+_i(t-1), S^-_i(t-1), \hat{\boldsymbol{w}}^{t}_i\right)$ \Comment{Sampling} \label{line:sampling}
            \EndFor
        \EndIf
        \If {$\theta = \text{optimization}$}
            \State $\boldsymbol{u}^t \gets \mathcal{O}\left(X, \hat{\boldsymbol{w}}^t, S^+(t - 1), \left\{S^-_i(t - 1)\right\}_{i=1}^m, z\right)$ \Comment{Optimization} \label{line:optimization}
        \EndIf
        \For {$i \in I$}
            \If {$\boldsymbol{u}^{t}_i \leq \boldsymbol{s}^+_i$ for any $\boldsymbol{s}^+_i \in S^+_i(t-1)$} \label{line:infer_label_1}
                \State $\ell^{t}_i \gets 1$
            \Else
                \State $\ell^{t}_i \gets \Omega_i(\boldsymbol{u}^{t}_i)$ \Comment{Oracle call} \label{line:oracle}
                \State $n_i \gets n_i + 1$
                \If {$\ell^{t}_i$ = 1}
                    \State $P^+_i(t) \gets P^+_i(t) \cup \{\boldsymbol{u}^{t}_i\}$
                \Else
                    \State $P^-_i(t) \gets P^-_i(t) \cup \{\boldsymbol{u}^{t}_i\}$
                \EndIf
            \EndIf
        \EndFor
        \If {$\theta = \text{optimization}$}
            \State $S^+(t) \gets S^+(t-1)$; $S^-(t) \gets S^-(t-1)$
            \If {$\ell^{t}_i = 1$ for every $i \in I$}
                \State $S^+(t) \gets S^+(t) \cup \{\boldsymbol{u}^t\}$
                \If {$z(\boldsymbol{u}^t) > L\!B$}
                    \State $L\!B = z(\boldsymbol{u}^t)$
                \EndIf
            \Else
                \State $S^-(t) \gets S^-(t) \cup \{\boldsymbol{u}^t\}$
            \EndIf
        \EndIf
    \EndFor
    \For {$i \in I$}
        \State $S^+_i(t) \gets S^+_i(t-1) \cup P^+_i(t)$; $S^-_i(t) \gets S^-_i(t-1) \cup P^-_i(t)$
    \EndFor
    \State $U\!B \gets \mathcal{B}\left(X, \left\{W_i\right\}_{i=1}^m, S^+(t), S^-(t), \left\{S^-_i(t)\right\}_{i=1}^m, z\right)$ \Comment{Bounding} \label{line:bounding}
\EndWhile
\State Get $\hat{\boldsymbol{x}} \in \arg\max\left\{z(\boldsymbol{x})\ |\ \boldsymbol{x} \in S^+(t)\right\}$
\For {$i \in I$}
    \State $\hat{\boldsymbol{w}}_i \gets \mathcal{S}(W_i, S^+_i(t), S^-_i(t))$
\EndFor
\State \textbf{return} $\hat{\boldsymbol{x}}$, $\left\{\hat{\boldsymbol{w}}_i\right\}_{i=1}^m$
\end{algorithmic}
\end{algorithm}

We can conceptually identify a few main steps inside the framework: the stopping criterion (line \ref{line:stopping}) is triggered by reaching the threshold on the relative gap or the maximum number of queries for any of the oracles. The linear separation step (line \ref{line:separation}) finds the surrogates weights for each unknown constraint by computing $\mathcal{S}(W_i, S^+_i(t-1), S^-_i(t-1))$; the sampling step (line \ref{line:sampling}) is the application of the chosen sampling strategy, i.e., $\mathcal{T}\left(X_i, S^+_i(t-1), S^-_i(t-1), \hat{\boldsymbol{w}}^{t}_i\right)$, to select an unlabeled point for each sub-problem; the optimization step (line \ref{line:optimization}) finds an optimal solution of the partially specified 0-1 IP model using the current surrogate knapsack constraints in place of the unknown ones. Both sampling and optimization are followed by an oracle calling phase (line \ref{line:oracle}). Each sampled point is submitted to the oracle corresponding to its unknown constraint. The solution of the optimization step is divided into as many sub-solutions as the number of unknown constraints. Each sub-solution is then submitted to the corresponding oracle. While each sampling strategy step always returns a sub-solution that has never been labeled before, the optimization step might yield a global solution such that some of the corresponding sub-solutions are positive and either they are already labeled or their label can be inferred from that of a previously labeled sub-solution without calling any oracle. We do this check in line \ref{line:infer_label_1}. The sets $S^+(t-1)$ and $S^-(t-1)$ are updated only using solutions of the optimization step, while for each $i \in I$ the sets $S^+_i(t-1)$ and $S^-_i(t-1)$ are updated using both the sub-solution coming from the $i$-th sampling strategy execution and the sub-solution obtained by restricting the global solution of the optimization step. The bounding step (line \ref{line:bounding}) computes a valid upper bound. To accomplish this task, we rely on a mixed-integer quadratic model, presented later in this section.

We remark that, since the sampling step returns unlabeled sub-solutions and the optimization step returns an unlabeled global solution with respect to the previous iterations, and because the set $X$ is finite, any instance of the framework ensures finite convergence to the optimal solution when permitting an unlimited number of oracle calls. Finally, we observe that, in principle, the framework admits modifications as, e.g., employing multiple sampling or optimization steps between any two consecutive linear separation steps, or selecting what sampled points to query at every time step $t$. \\

As mentioned in Section \ref{sec:BackgroundAndLiteratureReview}, a common choice for linear separation in active learning for binary classification is using an SVM. Specifically, we may solve the quadratic programming model
\begin{align*}
    \text{min} &\ \sum_{j=1}^n \omega_j^2 \\
    \text{s.t.} &\ \beta - \sum_{j=1}^n \omega_j \cdot \sigma^+_j \geq 1, & \boldsymbol{\sigma}^+ \in S^+_i(t), \\
    &\ \beta - \sum_{j=1}^n \omega_j \cdot \sigma^-_j \leq -1, & \boldsymbol{\sigma}^- \in S^-_i(t), \\
    &\ \omega_j \leq \beta, & j \in J \\
    &\ \boldsymbol{\omega} \geq 0,~\beta \in \mathbb{R}.
\end{align*}
and set $\hat{\boldsymbol{w}}_i := \boldsymbol{\omega}^*/\beta^*$ as surrogate weights. \\

For the sampling step, one can consider multiple existing approaches as well. Unfortunately, established SVM-related strategies like, e.g., max-min margin or ratio margin, require to train a linear classifier for every unlabeled point \cite{kremer2014active}. In general, most of the sampling strategies in the literature are incompatible with a combinatorial optimization setting, in which there is an exponential number of unlabeled examples to choose from. To the best of our knowledge, the only existing strategy that can be efficiently adapted to such a setting is simple margin (SIM) \cite{tong2001active}. It consists in selecting at every iteration $t$ the point that maximizes the uncertainty with respect to the classification provided by the current SVM, i.e., the unlabeled point closest to the decision boundary. The task can be accomplished by solving the following optimization problem:
\begin{subequations}
\begin{align}
    \text{min} &\ \left| 1 - \sum_{j=1}^n \hat{w}^{t}_{ij} \cdot \mu_j \right| \label{obj:SIM} \\
    \text{s.t.} &\ \sum_{j=1}^n \mu_j \cdot (1 - \sigma^+_j) \geq 1, & \boldsymbol{\sigma}^+ \in S^+_i(t), \label{constr:SIMFeasNGC} \\
    &\ \sum_{j=1}^n \mu_j \cdot \sigma^-_j \leq \sum_{j=1}^n \sigma^-_j - 1, &\boldsymbol{\sigma}^- \in S^-_i(t), \label{constr:SIMInfeasNGC} \\
    & \boldsymbol{\mu} \in X_i.\label{con:integrality}
\end{align}
\end{subequations}
The objective function \eqref{obj:SIM} minimizes the distance between the generic point $\boldsymbol{\mu} \in X_i$ and the hyperplane identified by $\hat{\boldsymbol{w}}_i$. No-good cuts \eqref{constr:SIMFeasNGC} and \eqref{constr:SIMInfeasNGC} are used to prevent finding a solution whose feasibility or infeasibility is implied by labeled points in the sets $S^+_i(t)$ and $S^-_i(t)$. We remark that, although integrality requirements \eqref{con:integrality} make the problem NP-hard in the general case, mixed-integer optimization models can be often solved effectively, e.g., by off-the-shelf discrete optimization solvers. \\

In order to compute the upper bound to the value of the best labeled feasible solution, we jointly perform convex separation and optimization. In particular, we find a polyhedron identified by $m$ hyperplanes that encloses all the labeled feasible points $S^+(t)$ but none of the labeled infeasible points $S^-(t)$. Simultaneously, we find a solution not already in $S^-(t)$ that maximizes the same objective function of Problem \eqref{model:01LinearProgramWithKnapsackConstraints}. The task can be accomplished by solving the following model:
{
\allowdisplaybreaks
\begin{subequations}
\label{model:Bounding}
    \begin{align}
        \max~ & z(\boldsymbol{x}) = \sum_{i=1}^m \sum_{j=1}^n v_{ij} \cdot x_{ij} \\
        \text{s.t.}~ & \sum_{j=1}^n w_{ij} \cdot x_{ij} \leq 1, & i \in I, \label{constr:BoundingKnapsackConstraints} \\
        & \sum_{j=1}^n w_{ij} \cdot \sigma^+_j \leq 1, & \boldsymbol{\sigma}^+ \in S^+_i(t),\ i \in I, \label{constr:BoundingFeasiblePoints} \\
        & \sum_{j=1}^n w_{ij} \cdot \sigma^-_j \geq 1, & \boldsymbol{\sigma}^- \in S^-_i(t), \ i \in I, \label{constr:BoundingInfeasiblePoints} \\
        & \sum_{j = 1}^n x_{ij} \cdot \sigma^-_j \leq \sum_{j=1}^n \sigma^-_j - 1, & \boldsymbol{\sigma}^- \in S^-_i(t),\ i \in I, \label{constr:BoundingNoGoodCuts} \\
        & \boldsymbol{x} \in X \subseteq \{0, 1\}^{m \times n}, \label{constr:BoundingCombinatorial} \\
        & \boldsymbol{w} \in W. \label{constr:BoundingParameterDomain}
    \end{align}
\end{subequations}
}
The quadratic constraints \eqref{constr:BoundingKnapsackConstraints} ensure that the optimal $\boldsymbol{x}^*$ belongs to the polyhedron identified by $\boldsymbol{w}^*$. Constraints \eqref{constr:BoundingFeasiblePoints} and \eqref{constr:BoundingInfeasiblePoints} impose that known sub-solutions are classified according to their label by their respective hyperplanes. More specifically, this implies that every point in $S^+(t)$ will belong to the polyhedron and every point in $S^-(t)$ will be outside or on the border. Finally, constraints \eqref{constr:BoundingNoGoodCuts} avoid that any sub-solution of $\boldsymbol{x}^*$ may be equal to a labeled sub-solution that makes invalid its relative unknown constraint. In particular, this constraints ensure that $\boldsymbol{x}^*$ is different from any infeasible point in $S^-(t)$.

The value of an optimal solution of Problem \eqref{model:Bounding} is an upper bound to the value of an optimal solution of Problem \eqref{model:01LinearProgramWithKnapsackConstraints}, as stated in the following Proposition and proved in Appendix \ref{sec:Proofs}.

\begin{proposition}
\label{prop:Bounding}
For any optimal solution $(\boldsymbol{x}^*$, $\boldsymbol{w}^*)$ of Problem \eqref{model:Bounding} and for any optimal solution $\boldsymbol{x}^{\circledast}$ of Problem \eqref{model:01LinearProgramWithKnapsackConstraints}, the value of $\boldsymbol{x}^*$ is greater of equal than the value of $\boldsymbol{x}^{\circledast}$, i.e., $z(\boldsymbol{x}^*) \geq z(\boldsymbol{x}^{\circledast})$.
\end{proposition}

\subsection{Closest Cutting Plane Sampling Strategy}
\label{subsec:VersionSpaceAndCLosestCuttingPlaneSamplingStrategy}
For the pair of sets $S^+_i(t)$ and $S^-_i(t)$ of labeled sub-solutions for the unknown constraint $i$ at time step $t$, the current version space is defined as the set
\begin{align*}
    V_i(t) & = \{\boldsymbol{\omega} \in W_i\ |\ \boldsymbol{\omega} \cdot \boldsymbol{\sigma}^+ \leq 1\ \text{for all}\ \boldsymbol{\sigma}^+ \in S^+_i(t)\text{,}\ \boldsymbol{\omega} \cdot \boldsymbol{\sigma}^- > 1\ \text{for all}\ \boldsymbol{\sigma}^- \in S^-_i(t) \}\text{.}
\end{align*}
Let us also define $S^+_i(\infty) = \{\boldsymbol{\chi} \in X_i\ |\ \bar{\boldsymbol{w}}_i \cdot \boldsymbol{\chi} \leq 1\}$ and $S^-_i(\infty) = \{\boldsymbol{\chi} \in X_i\ |\ \bar{\boldsymbol{w}}_i \cdot \boldsymbol{\chi} > 1\}$, as the sets of all sub-solutions that satisfy and violate the unknown constraint $i$, respectively. Every time we update $S^+_i(t)$ and $S^-_i(t)$ with new labeled points to obtain $S^+_i(t+1)$ and $S^-_i(t+1)$, the next version space $V_i(t+1)$ will be defined by additional inequalities with respect to $V_i(t)$. The sequence of version spaces converges (in a finite but potentially exponential number of steps) to the set of surrogate weights whose corresponding surrogate constraint classifies every point in $X_i$ on the same side as the unknown knapsack constraint $\bar{\boldsymbol{w}}_i \cdot \boldsymbol{\chi} \leq 1$, namely,
\begin{align}
    V_i(\infty) & = \{\boldsymbol{\omega} \in W_i\ |\ \boldsymbol{\omega} \cdot \boldsymbol{\chi} \leq 1\ \text{if and only if}\ \bar{\boldsymbol{w}}_i \cdot \boldsymbol{\chi} \leq 1\ \text{for all}\ \boldsymbol{\chi} \in X_i\} \nonumber\\
    & = \{\boldsymbol{\omega} \in W_i\ |\ {\boldsymbol{\omega}} \cdot \boldsymbol{\sigma}^+ \leq 1\ \text{for all}\ \boldsymbol{\sigma}^+ \in S^+_i(\infty)\text{,}\ \\
    & \ \ \ \ \ \ \ \ \ \ \ \ \ \ \ \ \ \ \ \! \! \boldsymbol{\omega} \cdot \boldsymbol{\sigma}^- > 1\ \text{for all}\ \boldsymbol{\sigma}^- \in S^-_i(\infty) \}\text{.} \label{def:V_inf}
\end{align}
Intuitively, if the vector of surrogate constraints $\hat{\boldsymbol{w}}_i$ is sufficiently close to $V_i(\infty)$, then sampling an unlabeled sub-solution $\boldsymbol{\mu}$ from $X_i$ whose corresponding cut $\boldsymbol{\omega} \cdot \boldsymbol{\mu} = 1$ is close to $\hat{\boldsymbol{w}}_i$ can have the effect of significantly reducing the size of the version space, hence accelerating the convergence. Following this intuition, we propose a sampling strategy alternative to SIM, named closest cutting plane (CUT). Given $\hat{\boldsymbol{w}}_i$ in the closure $\bar{V}_i(t)$ of the current version space, CUT finds an unlabeled point $\boldsymbol{\mu} \in X_i$ whose corresponding hyperplane $h_{\boldsymbol{\mu}} = \{\boldsymbol{\omega} \in \mathbb{R}^n\ |\ \boldsymbol{\omega} \cdot \boldsymbol{\mu} = 1\}$ has minimum Euclidean distance from $\hat{\boldsymbol{w}}_i$. To accomplish this task, we solve the following mixed-integer quadratic program (MIQP): 
\begin{subequations}\label{model:CUT}
\begin{align}
    \text{min} &\ \sum_{j=1}^n (p_j - \hat{w}^t_{ij})^2 \label{obj:CUT} \\
    \text{s.t.} &\ \sum_{j=1}^n p_j \cdot \mu_j = 1, \label{constr:CUTQuadConstr} \\
    &\ \sum_{j=1}^n \mu_j \cdot (1 - s^+_j) \geq 1,&&\boldsymbol{s}^+ \in S^+_i(t), \label{constr:CUTFeasNGC} \\
    &\ \sum_{j=1}^n \mu_j \cdot s^-_j \leq \sum_{j=1}^n s^-_j - 1,&&\boldsymbol{s}^- \in S^-_i(t), \label{constr:CUTInfeasNGC} \\
    &\ \boldsymbol{\mu} \in X_i,~\boldsymbol{p} \in \mathbb{R}^n.
\end{align}
\end{subequations}
The distance between $\hat{\boldsymbol{w}}_i$ and the hyperplane $h_{\boldsymbol{\mu}}$ is modeled as the distance between $\hat{\boldsymbol{w}}_i$ and its projection $\boldsymbol{p}$ on $h_{\boldsymbol{\mu}}$, which is minimized by the objective function \eqref{obj:CUT}. The quadratic constraint \eqref{constr:CUTQuadConstr} ensures that $\boldsymbol{p}$ belongs to $h_{\boldsymbol{\mu}}$. No-good cuts \eqref{constr:CUTFeasNGC} and \eqref{constr:CUTInfeasNGC} are used exactly as in SIM.

\subsection{Approximation of a Separation Oracle and the Resulting Linear Separation Algorithm}

In principle, given a vector of surrogate weights in $V_i(\infty)$ for every $i \in I$, one can solve the original combinatorial optimization problem \eqref{model:01LinearProgramWithKnapsackConstraints} by replacing the unknown constraints with the surrogate constraints. Therefore, in order to solve \eqref{model:01LinearProgramWithKnapsackConstraints}, one can try to find for each $i \in I$ a point in the convex set $V_i(\infty)$ using a membership oracle of the constraint $\bar{\boldsymbol{w}}_i \cdot \boldsymbol{\chi} \leq 1$ with respect to $X_i$. It is well known that one can efficiently find a point in a well-bounded convex body $K$ with a separation oracle. Specifically, assuming that a compact convex set $K \subseteq \mathbb{R}^n$ is contained in the Euclidean ball of radius $R$ centered at the origin, there exist algorithms using $O(n\log(nR/\epsilon))$ calls to the separation oracle to either find a point in $K$ or prove that $K$ does not contain any Euclidean ball of radius $\epsilon$ \cite{dadush2023simple,jiang2020improved,lee2015faster}. However, such algorithms are not directly applicable in our case because we do not have direct access to a separation oracle of $V_i(\infty)$. On the other hand, given a weak separation oracle of a convex body $K$ (that finds an almost violated linear inequality valid for $K$), there exist algorithms that can find a point almost in $K$ in oracle-polynomial time \cite{grotschel2012geometric}. To hopefully find a point in $V_i(\infty)$ (or at least get a good approximation of $V_i(\infty)$), we propose to construct an approximate separation oracle of $V_i(\infty)$ using the membership oracle of constraint $i$ in \eqref{constr:KnapsackConstraints}, that characterizes the set $S^+_i(\infty)$ of all the sub-solutions satisfying the constraint, and replace the separation oracle in existing convex optimization algorithms by the constructed approximate separation oracle. In particular, we adapt the algorithm proposed in \cite{dadush2023simple} as our choice of the convex optimization algorithm in the oracle model.

The original version of the algorithm presented in \cite{dadush2023simple} is intended to minimize a convex function $f$ over a full-dimensional convex body $K\subseteq \mathbb{R}^n$. Specifically, it generates a sequence of points in $\mathbb{R}^n$ that converges to a minimum inside $K$, under the assumption that $K$ is contained inside a ball of radius $R$ centered in the origin. Such a set $K$ is given implicitly by means of a separation oracle, which for every point $\boldsymbol{\omega} \in \mathbb{R}^n$, either states that $\boldsymbol{\omega} \in K$ or returns an inequality $\boldsymbol{\alpha} \cdot \boldsymbol{\omega} \leq \beta$ valid for $K$ but violated by $\boldsymbol{\omega}$. To address the solution of Problem \eqref{model:01LinearProgramWithKnapsackConstraints}, we assume $f$ to be constant and we let $K$ be the set $V_i(\infty)$ defined in Section \ref{subsec:VersionSpaceAndCLosestCuttingPlaneSamplingStrategy}.
To approximate a separation oracle for $K$ by using the membership oracle at our disposal, we collect a valid inequality for $K$ every time we label a point
regardless of whether the current surrogate weights $\hat{\boldsymbol{w}}_i$ violate the inequality or not. For each feasible point $\boldsymbol{\sigma}^+ \in S^+_i(\infty)$, the inequality is $\boldsymbol{\omega} \cdot \boldsymbol{\sigma}^+ \leq 1$. For each infeasible point $\boldsymbol{\sigma}^- \in S^-_i(\infty)$, the inequality is $-\boldsymbol{\omega} \cdot \boldsymbol{\sigma}^- \leq -1$. This approximation is designed to be effective when using CUT 
as a sampling strategy, as the intuition is that if we can obtain objective value $0$ in \eqref{model:CUT}, then the valid inequality corresponding to the sampled point separates $\hat{\boldsymbol{w}}_i$ from $K$. \\

As in \cite{dadush2023simple}, let us define the norm $\|(\boldsymbol{x}, y)\| = \sqrt{2} \|(x_1 / R, \ldots, x_n / R, 1)\|_2$ and the potential $\Phi(\boldsymbol{\alpha}, \beta) = \frac{1}{4}\|(R\boldsymbol{\alpha}, \beta)\|_2^2$, with $R = \sqrt{n}$, the diameter of $[0, 1]^n$. The pseudo-code for the resulting algorithm, that we call SEP, is given in Algorithm \ref{alg:AlternativeSeparator}. The weight domain $W_i$ in input is a polyhedron identified by a finite number of linear inequalities, that we can define as
\begin{equation}
    W_i = \{\boldsymbol{\omega} \in [0, 1]^n\ |\ \boldsymbol{a} \cdot \boldsymbol{\omega} \leq b\ \ \forall (\boldsymbol{a}, b) \in \Delta_i\}\text{,}
\end{equation}
where $\Delta_i \subseteq \mathbb{R}^n \times \mathbb{R}$ is finite. The set $\Lambda$ collects all currently available valid inequalities for $K$, which are the same that describe the current version space $V_i(t) \supseteq K$ as defined in Section \ref{subsec:VersionSpaceAndCLosestCuttingPlaneSamplingStrategy} (including those that define $W_i$). Each inequality $\boldsymbol{\alpha} \cdot \boldsymbol{\omega} \leq \beta$ is encoded as a pair $(\boldsymbol{\alpha}, \beta)$ and divided by the norm $\|(\boldsymbol{\alpha}, \beta)\|$ of the pair. 
\begin{algorithm}[t]
\caption{Alternative linear separator (SEP)}\label{alg:AlternativeSeparator}
\begin{algorithmic}[1]
\State \textbf{Input:} weight domain $W_i$, iteration counter $t$, sets $S^+_i(t)$ and $S^-_i(t)$ of feasible and infeasible points respectively
\State $\Lambda \gets \{(\boldsymbol{a}, b) / \| (\boldsymbol{a}, b) \|_*\ |\ (\boldsymbol{a}, b) \in \Delta\}$
\For {$\boldsymbol{\sigma}^+ \in S^+_i(t)$}
    \State $\Lambda \gets \Lambda \cup \{(\boldsymbol{\sigma}^+, 1) / \| (\boldsymbol{\sigma}^+, 1) \|_*\}$
\EndFor
\For {$\boldsymbol{\sigma}^- \in S^-_i(t)$}
    \State $\Lambda \gets \Lambda \cup \{(-\boldsymbol{\sigma}^-, -1) / \| (-\boldsymbol{\sigma}^-, -1) \|_*\}$
\EndFor
\State Get a point $\boldsymbol{\gamma}^t \in \arg\min\{\Phi(\gamma)\ |\ \gamma \in \text{conv}(\Lambda)\}$ \label{line:MinPotential}
\State $\hat{\boldsymbol{w}}^t_i \gets - \left( \nabla\Phi(\gamma^t)_1, \ldots, \nabla\Phi(\gamma^t)_d \right) / \nabla\Phi(\gamma^t)_{d + 1}$ \label{line:ComputeSurrogateWeights}
\State \textbf{return} $\hat{\boldsymbol{w}}^t_i$
\end{algorithmic}
\end{algorithm}

 The original algorithm from \cite{dadush2023simple} guarantees a convergence rate towards a vector in $K$. However, in our case, no guarantee can be given regarding the convergence of $\hat{\boldsymbol{w}}^t_i$ towards a vector in $K = V_i(\infty)$, because we are approximating the separation oracle required by the original algorithm using a membership oracle. Therefore, we focus on the empirical performance of the algorithm rather than on theoretical guarantees.

\section{Test Problems}
\label{sec:TestProblems}

In this Section, we give the details of the problems on which we test the ISEO framework and compare different combinations of separation methods and sampling strategies. \\

\noindent
{\bf Knapsack Problem.} Knapsack problems are a family of massively studied problems whose origins date back to the late 19th century \cite{mathews1896partition}, whilst the first known algorithmic studies were conducted in the 1950s \cite{dantzig1957discrete,bellman1957comment}. For an overview of the problem variants and details about a wide variety of solution approaches, the reader may refer to a recent survey in two parts \cite{cacchiani2022knapsackA,cacchiani2022knapsackB}. Here, we are specifically interested in the ancestor of all Knapsack Problems, i.e. the 0-1 Knapsack Problem. In this problem, a selection of items has to be packed inside a backpack, respecting its capacity. Each item is characterized by a weight and a value. The goal is to maximize the total value of the items inside the backpack. The problem is NP-hard and has pseudo-polynomial time complexity with respect to the capacity value. \\

\noindent The 0 - 1 Knapsack Problem requires to solve the following model:
\begin{subequations}
    \allowdisplaybreaks
    \begin{align*}
        \max~ & z(\boldsymbol{x}) := \sum_{i=1}^n v_j \cdot x_j \\
        \text{s.t.}~ & \sum_{j=1}^n \bar{w}_{j} \cdot x_{j} \leq 1, \\
        & \boldsymbol{x} \in \{0, 1\}^n.
    \end{align*}
\end{subequations}
With reference to Problem \eqref{model:01LinearProgramWithKnapsackConstraints}, the unknown constraint is also the only linear constraint, and there are no additional known constraints. \\

\noindent
{\bf College Study Plan Problem.}
In order to have a test problem with a single unknown knapsack constraint but also additional known constraints, we introduce the College Study Plan Problem (CSPP), that we obtained as a variant of the precedence constrained knapsack problem proposed in \cite{aslan2023precedence}. In this 0-1 combinatorial optimization problem, a prospective college student needs to design their study plan in order to maximize the number of credits acquired respecting an estimated time budget. Each course yields a predefined number of credits and requires an effort in terms of study hours given by a certain fixed number and an amount of hours for independent study that depends on the particular student. Given a possible study plan, the student needs to decide whether or not taking the courses specified in the plan is feasible with respect to the time budget. We assume that an unknown knapsack constraint models the satisfaction of the time budget. Additional known constraints model the dependencies between the available courses present in the course catalog. Any course $i$ can be included in a study plan only if at least one course from each of a sequence of prerequisite sets $P^i_1$, $P^i_2$, $\ldots$, $P^i_{N^i}$ is included in the plan. Every course $i$ also has a set $C^i$ of corequisites (courses that must be taken if and only if course $i$ is taken) and a set $A^i$ of alternatives (courses that cannot be taken if $i$ is taken). Reasonable assumptions avoid inconsistency of the dependencies, namely: (a) $\bigcup_{k=1}^{N^i} P^i_k$, $C^i$ and $A^i$ are pairwise disjoint sets for every $i \in \{1, \ldots, d\}$; (b) if $i \in \bigcup_{k=1}^{N^j} P^j_k$ then $j \notin \bigcup_{k=1}^{N^i} P^i_k$, for every $i, j \in \{1, \ldots, d\}$; (c) $i \in C^j$ if and only if $j \in C^i$ for every $i, j \in \{1, \ldots, d\}$; (d) $i \in A^j$ if and only if $j \in A^i$ for every $i, j \in \{1, \ldots, d\}$. The goal is to solve the following 0-1 integer linear programming model:
\begin{align*}
    \max~ & \sum_{i=1}^d v_i \cdot x_i \\
    \text{s.t.}~ & \sum_{i=1}^d \bar{w}_i \cdot x_i \leq 1, \\
    & x_i \leq \sum_{j \in P^i_k} x_j, &  &k \in \{1, \ldots, N^i\}, \\
    & x_i = x_j, && j \in C^i, \\
    & x_i + x_j \leq 1, && j \in A^i, \\
    & x_i \in \{0, 1\}, && i \in \{1, \ldots, d\}.
\end{align*}
Here, each reward $v_i \geq 0$ represents the number of credits associated to course $i$. Each weight in the knapsack constraint can be modeled as $\bar{w}_i = \min\{\eta_i + \xi_i, 1\}$, where $\eta_i \geq 0$ represents the known number of course hours $i$ and $\xi_i \geq 0$ the unknown student-dependent component, assuming that both are normalized by dividing them by the unknown total time budget. \\

\noindent
{\bf Generalized Assignment Problem.}
The Generalized Assignment Problem \cite{cattrysse1992survey,oncan2007survey} (GAP) is a very well studied problem in its classical formulation, which appeared for the first time in the 1970s \cite{ross1975branch}. We are given a set of agents, as well as a set of jobs to be completed. Each job can be assigned to no more that one agent. Each assignment of a job to an agent is characterized by a reward and a processing time. The goal is to find an assignment of the jobs to the agents whose total reward is maximum, respecting a time budget constraint on each agent. The problem is also a generalization of the 0-1 Knapsack Problem, in which there are multiple knapsacks and the value of each item depends on the knapsack it is assigned to. The problem is NP-hard \cite{fisher1986multiplier} and the associated decision problem is NP-complete \cite{martello1990knapsack}. \\

\noindent The 0-1 integer programming model for the Generalized Assignment Problem is the following:
\begin{subequations}
\label{model:GeneralizedAssignment}
    \begin{align}
        \max~ & z(\boldsymbol{x}) = \sum_{i=1}^m \sum_{j=1}^n v_{ij} \cdot x_{ij} \\
        \text{s.t.}~ & \sum_{j=1}^n \bar{w}_{ij} \cdot x_{ij} \leq 1, & i \in I, \label{constr:GeneralizedAssignmentKnapsackConstraints} \\
        & \sum_{i = 1}^m x_{ij} \leq 1, & j \in J, \label{constr:GeneralizedAssignmentPartitioningConstraints} \\
        & \boldsymbol{x} \in \{0, 1\}^{m \times n}.
    \end{align}
\end{subequations}
In our interactive setting, the model contains both the $m$ unknown constraints \eqref{constr:GeneralizedAssignmentKnapsackConstraints} and additional known partitioning constraints \eqref{constr:GeneralizedAssignmentPartitioningConstraints}. For such an interactive version of the problem, an example of a real application is given in Appendix \ref{apd:application}.

\section{Data Collection and Generation}
\label{sec:DataCollectionAndGeneration}

In the following, we show how we generated the test instances for the Knapsack Problem and the College Study Plan Problem, and how we collected the test instances of the Generalized Assignment Problem. \\

\noindent
{\bf Knapsack Problem.}
We applied a Python re-implementation of a well known knapsack generator \cite{pisinger1999core} to obtain test instances for the Knapsack Problem (KNAP) with 60 items, with rewards and weights of three types: uncorrelated (KNAP-U), weakly correlated (KNAP-W) and strongly correlated (KNAP-S). For each type, we generated 30 instances, each with items having integer weights in the range $[1, 10000]$. For each instance number $h \in \{1, \ldots, 30\}$, the corresponding capacity is equal to $h$ multiplied by the sum of the weights and divided by 31. \\

\noindent
{\bf College Study Plan Problem.}
For CSPP, inspired by the experimental setting of \cite{aslan2023precedence}, we designed a set of instances according to the structure of the undergraduate courses in Mathematics at the Massachusetts Institute of Technology (MIT), whose list can be found here:
\begin{center}
    \hyperlink{http://student.mit.edu/catalog/m18a.html}{http://student.mit.edu/catalog/m18a.html}.
\end{center}
Differently from \cite{aslan2023precedence}, we took into account the entire list and we also considered every dependency inside other areas (Physics, Computer Science, Biology, etc.). For every course, the credits obtainable for passing the corresponding exam are specified as a triple $(a_i, b_i, c_i)$, where the three values represent credits assigned for lectures and recitations, for laboratory, design, or field work, and for outside (independent) preparation, respectively. Each credit corresponds to 14 hours of activity, but while this equivalence is exact for $a_i$ and $b_i$, the hours actually dedicated to outside preparation vary from student to student. We operated marginal adjustments to the original data, mainly consisting in merging homologous courses, and we obtained a dataset with 150 courses whose structure of dependencies respect the definition of the college study plan problem. We generated 30 test instances, where the parameters related to each course $i$ are computed according to the following specifications from \cite{aslan2023precedence} ($B$ denotes the total hours budget equal to 1700, while $\mathcal{N}_\text{T}$ denotes the truncated normal distribution supported on $[0,B]$):
\begin{enumerate}
    \item $\eta_i = 14 \cdot (a_i + b_i) / B$;
    \item $\xi_i \sim \mathcal{N}_\text{T}(\mu_i, \sigma_i^2) / B$, where $\mu_i = 14\cdot c_i$ and $\sigma_i \sim U[1, 28]$;
    \item $v_i = a_i + b_i + c_i$.
\end{enumerate}

\noindent
{\bf Generalized Assignment Problem.}
The experiments are based on instances of the Generalized Assignment Problem downloaded from OR-Library \cite{beasley1990or}. In particular, we used the instances contained in the files marked as ``gap\_1'' through ``gap\_12''. The number $m$ of agents is 5, 8 or 10. The number $n$ of jobs is 3, 4, 5 or 6 times the number or agents. For each pair of $m$ and $n$, there are 5 different instances.

\section{Experimental Setting and Results}
\label{sec:ExperimentalSettingAndResults}

We implemented the ISEO algorithms using Python 3.11.4. The experiments were conducted on a Linux machine with two Intel XEON Platinum 8575C processors. The optimization models were solved by using the Gurobi solver with a limit of 2 threads on each optimization. A global time limit of 500,000 s (approximately 5.8 days) was imposed for each problem instance tested. A portion of the models for SIM and CUT to be solved could require an excessive amount of time to reach optimality, therefore we applied a node limit equal to 50,000 for the Gurobi branch-and-bound algorithm. \\

The algorithm was executed on the test instances described in Section \ref{sec:DataCollectionAndGeneration} with the following choice of parameters: the oracle call limit $N$ is equal to 2,000; the initial sets of labeled points are chosen as $S^+(0) = \{\boldsymbol{0}\}$ and $S^-(0) = \{\boldsymbol{1}\}$; the surrogate domains are $W_i = [0, 1]^n$ for every $i \in I$; and, finally, the gap threshold $thr$ is set to 0.01. \\

Tables \ref{tab:kp_1} - \ref{tab:gap_2} show the results of the experiments. In particular, Tables \ref{tab:kp_1} and \ref{tab:kp_2} refer to experiments on problems with a single unknown constraint, i.e. the Knapsack Problem and the College Study Plan Problem. The first column (Type) indicates the specific subset of instances. Tables \ref{tab:gap_1} and \ref{tab:gap_2} are relative to the experiments on the Generalized Assignment Problem. The first 2 columns show the number of agents ($m$) and jobs ($n$). The remaining columns of the four tables have the following meaning:

\begin{enumerate}
    \item [$\cdot$] Separ. -- separation algorithm of choice;
    \item [$\cdot$] Sampl. -- sampling strategy of choice;
    \item [$\cdot$] \# Opt. -- number of instances solved to optimality;
    \item [$\cdot$] \# Thres. -- number of instances for which the gap threshold was reached;
    \item [$\cdot$] \# Feas. -- number of instances for which an additional optimization step returned a feasible solution. The optimization step is executed on the original instance after replacing the unknown constraints with the approximate constraints returned by the ISEO algorithm;
    \item [$\cdot$] Gap (\%) -- average final gap between upper bound and lower bound (value of the best labeled feasible solution), computed as $100 * (UB - LB) / LB$;
    \item [$\cdot$] Error (\%) -- average final error between the best labeled feasible solution and the actual optimal solution of the instance, computed as $100 * (z(x^*) - z(\hat{x})) / z(x^*)$;
    \item [$\cdot$] \# Calls -- average total number of oracle calls, summing up the number of calls over all the oracles;
    \item [$\cdot$] \# Iters -- average number of iterations;
    \item [$\cdot$] Time (s) -- running time fo the algorithm;
    \item [$\cdot$] \# Calls to Thres. -- average number of oracle calls required to reach the gap threshold, summing up the number of calls over all the oracles (computed over the instances that reached the gap threshold);
    \item [$\cdot$] \# Iters to Thres. -- average number of iterations required to reach the gap threshold (computed over the instances that reached the gap threshold);
    \item [$\cdot$] Time to Thres. (s) -- average running time required to reach the threshold (computed over the instances that reached the gap threshold);
    \item [$\cdot$] \# Calls to Opt. -- average number of oracle calls required to solve the instance to optimality, summing up the number of calls over all the oracles (computed over the instances that reached an optimal solution);
    \item [$\cdot$] \# Iters to Opt. -- average number of iterations required to solve the instance to optimality (computed over the instances that reached an optimal solution);
    \item [$\cdot$] Time to Opt. (s) -- average running time required to solve the instance to optimality (computed over the instances that reached an optimal solution).
\end{enumerate}
\begin{table}[tb!]
\caption{Results of the experiments with instances of the Knapsack Problem and the College Study Plan Problem -- part 1.}
\centering 
\scriptsize
\begin{tabular}{c|l|l|r|r|r|r|r|r|r|r}
Type & Separ. & Sampl. & \# Opt. & \# Thres. & \# Feas. & Gap (\%) & Error (\%) & \# Calls & \# Iters & Time (s) \\ 
\hline \multirow{4}{*}{KNAP-U}
& SEP & CUT & 30/30 & 6/30 & 6/30 & 37.655 &     - & 1908.5 &  954.2 & 178336.73 \\
& SEP & SIM &  5/30 & 3/30 & 3/30 & 76.191 & 5.984 & 1949.8 &  974.9 &   1391.46 \\
& SVM & CUT & 26/30 & 4/30 & 4/30 & 41.390 & 0.014 & 2000.0 & 1000.0 & 160384.62 \\
& SVM & SIM &  4/30 & 3/30 & 3/30 & 69.923 & 7.492 & 2000.0 & 1000.3 &   7124.16 \\
\hline \multirow{4}{*}{KNAP-W}
& SEP & CUT & 28/30 & 0/30 & 0/30 & 152.127 & 0.006 & 2000.0 & 1000.0 & 191069.56 \\
& SEP & SIM &  0/30 & 0/30 & 0/30 & 214.300 & 6.653 & 2000.0 & 1000.0 &    1658.10 \\
& SVM & CUT & 28/30 & 0/30 & 0/30 & 162.489 & 0.009 & 2000.0 & 1000.0 & 186140.92 \\
& SVM & SIM &  0/30 & 0/30 & 0/30 & 192.981 & 6.737 & 2000.0 & 1000.1 &   7887.27 \\
\hline \multirow{4}{*}{KNAP-S}
& SEP & CUT &  0/30 & 0/30 & 0/30 & 156.447 & 0.049 & 2000.0 & 1000.0 &  198645.30 \\
& SEP & SIM &  0/30 & 0/30 & 0/30 & 220.291 & 0.234 & 2000.0 & 1000.0 &   2304.35 \\
& SVM & CUT &  0/30 & 0/30 & 0/30 & 289.137 & 0.081 & 2000.0 & 1000.0 & 203841.18 \\
& SVM & SIM &  0/30 & 0/30 & 0/30 & 199.856 & 0.197 & 2000.0 & 1000.1 &   7603.54 \\
\hline \multirow{4}{*}{CSPP}
& SEP & CUT & 6/30 & 0/30 & 0/30 & 1016.577 & 0.222 & 2000.0 & 1998.0 & 247593.64 \\
& SEP & SIM & 0/30 & 0/30 & 0/30 & 6916.667 & 82.324 & 2000.0 & 1998.0 & 3987.16 \\
& SVM & CUT & 7/30 & 0/30 & 0/30 & 1068.713 & 1.916 & 2000.0 & 1985.0 & 283439.67 \\
& SVM & SIM & 0/30 & 0/30 & 0/30 & 1351.949 & 14.529 & 2000.0 & 1998.0 & 3063.74 \\
\hline
\multicolumn{11}{l}{At each row -- column 4 to last -- statistics averaged over the specific set of instances are reported.}
\label{tab:kp_1}
\end{tabular}
\end{table}
\begin{table}[tb!]
\caption{Results of the experiments with instances of the Knapsack Problem and the College Study Plan Problem -- part 2.}
\centering 
\scriptsize
\begin{tabular}{c|l|l|r|r|r|r|r|r}
Type & Separ. & Sampl. & \makecell{\# Calls to\\Thres.} & \makecell{\# Iters to\\Thres.} & \makecell{Time to\\Thres. (s)} & \makecell{\# Calls to\\Opt.} & \makecell{\# Iters to\\Opt.} & \makecell{Time to\\Opt. (s)} \\
\hline \multirow{4}{*}{KNAP-U}
& SEP & CUT & 568.3 & 283.7 & 10239.77 & 621.80 & 310.40 & 18432.80 \\
& SEP & SIM & 224.3 & 111.7 & 14.26 & 548.60 & 273.8 & 152.67 \\
& SVM & CUT & 507.5 & 253.3 & 34233.89 & 826.4 & 412.7 & 39545.41 \\
& SVM & SIM & 23.0 & 11.0 & 0.43 & 262.8 & 131.3 & 64.81 \\
\hline \multirow{4}{*}{KNAP-W}
& SEP & CUT & - & - & - & 1267.6 & 633.3 & 84958.00 \\
& SEP & SIM & - & - & - & - & - & - \\
& SVM & CUT & - & - & - & 1339.1 & 669.0 & 92720.09 \\
& SVM & SIM & - & - & - & - & - & - \\
\hline \multirow{4}{*}{KNAP-S}
& SEP & CUT & - & - & - & - & - & - \\
& SEP & SIM & - & - & - & - & - & - \\
& SVM & CUT & - & - & - & - & - & - \\
& SVM & SIM & - & - & - & - & - & - \\
\hline \multirow{4}{*}{CSPP}
& SEP & CUT & - & - & - & 1479.7 & 1477.7 & 154587.88 \\
& SEP & SIM & - & - & - & - & - & - \\
& SVM & CUT & - & - & - & 1575.6 & 1560.6 & 201258.11 \\
& SVM & SIM & - & - & - & - & - & - \\
\hline
\multicolumn{9}{l}{At each row -- column 4 to last -- statistics averaged over the specific set of instances are reported.}
\label{tab:kp_2}
\end{tabular}
\end{table}

\begin{table}[tb!]
\caption{Results of the experiments with instances of the Generalized Assignment Problem -- part 1.}
\centering 
\scriptsize
\begin{tabular}{c|c|l|l|r|r|r|r|r|r|r|r}
$m$ & $n$ & Separ. & Sampl. & \# Opt. & \# Thres. & \# Feas. & Gap(\%) & Error (\%) & \# Calls & \# Iters & Time (s) \\ 
\hline \multirow{4}{*}{5} & \multirow{4}{*}{15} 
  & SEP & CUT & 5/5 & 5/5 & 5/5 &     - &     - &  1404.8 &  266.4 &    544.82 \\
& & SEP & SIM & 5/5 & 5/5 & 5/5 &     - &     - &  1931.2 &  366.2 &    435.69 \\
& & SVM & CUT & 5/5 & 5/5 & 5/5 &     - &     - &  1773.6 &  323.6 &    842.78 \\
& & SVM & SIM & 5/5 & 5/5 & 5/5 &     - &     - &  1886.8 & 1059.0 &  13178.48 \\
\hline \multirow{4}{*}{5} & \multirow{4}{*}{20}
  & SEP & CUT & 5/5 & 5/5 & 5/5 &     - &     - &  1515.8 &  254.2 &   3991.95 \\
& & SEP & SIM & 5/5 & 5/5 & 5/5 & 0.187 &     - &  9142.6 & 1707.6 &  16258.52 \\
& & SVM & CUT & 5/5 & 5/5 & 5/5 &     - &     - &  2369.2 &  405.2 &   6326.38 \\
& & SVM & SIM & 5/5 & 5/5 & 5/5 & 0.370 &     - &  9004.2 & 1683.0 &  17012.24 \\
\hline \multirow{4}{*}{5} & \multirow{4}{*}{25}
  & SEP & CUT & 5/5 & 5/5 & 5/5 & 0.035 &     - &  3957.6 &  697.6 &  32623.76 \\
& & SEP & SIM & 4/5 & 3/5 & 3/5 & 1.445 & 0.071 &  9524.0 & 1717.2 &  15883.95 \\
& & SVM & CUT & 5/5 & 5/5 & 5/5 &     - &     - &  2904.0 &  482.0 &  12133.27 \\
& & SVM & SIM & 5/5 & 1/5 & 1/5 & 1.194 &     - &  9452.6 & 1703.4 &  15993.39 \\
\hline \multirow{4}{*}{5} & \multirow{4}{*}{30}
  & SEP & CUT & 5/5 & 5/5 & 5/5 & 0.151 &     - &  4618.6 &  797.0 &   76176.00 \\
& & SEP & SIM & 1/5 & 0/5 & 0/5 & 3.576 & 0.461 &  9448.8 & 1662.8 &  13561.13 \\
& & SVM & CUT & 5/5 & 5/5 & 5/5 & 0.186 &     - &  5496.4 &  933.6 & 115284.77 \\
& & SVM & SIM & 2/5 & 0/5 & 0/5 & 2.966 & 0.181 &  9416.6 & 1670.0 &  16284.04 \\
\hline \multirow{4}{*}{8} & \multirow{4}{*}{24}
  & SEP & CUT & 5/5 & 5/5 & 5/5 &     - &     - &  5779.0 &  678.8 &  28186.97 \\
& & SEP & SIM & 5/5 & 5/5 & 5/5 &     - &     - & 10107.0 & 1205.4 &  12082.00 \\
& & SVM & CUT & 5/5 & 5/5 & 5/5 &     - &     - &  9132.2 & 1083.2 &  40916.83 \\
& & SVM & SIM & 5/5 & 5/5 & 5/5 &     - &     - &  9085.8 & 1078.4 &  10279.21 \\
\hline \multirow{4}{*}{8} & \multirow{4}{*}{32}
  & SEP & CUT & 5/5 & 5/5 & 5/5 & 0.053 &     - &  8758.4 & 1034.4 & 226832.13 \\
& & SEP & SIM & 5/5 & 1/5 & 1/5 & 1.193 &     - & 15609.4 & 1845.0 &  24290.42 \\
& & SVM & CUT & 5/5 & 5/5 & 5/5 & 0.238 &     - & 12629.4 & 1487.6 & 198461.04 \\
& & SVM & SIM & 4/5 & 0/5 & 0/5 & 1.377 & 0.026 & 15566.6 & 1842.4 &  24881.58 \\
\hline \multirow{4}{*}{8} & \multirow{4}{*}{40}
  & SEP & CUT & 5/5 & 5/5 & 5/5 & 0.106 &     - &  8231.0 &  940.4 & 365948.16 \\
& & SEP & SIM & 2/5 & 2/5 & 2/5 & 1.607 & 0.169 & 15200.2 & 1758.4 &  19429.76 \\
& & SVM & CUT & 5/5 & 5/5 & 5/5 & 0.380 &     - & 12715.2 & 1466.6 & 411785.08 \\
& & SVM & SIM & 4/5 & 2/5 & 2/5 & 1.355 & 0.148 & 15220.0 & 1765.2 &  51808.54 \\
\hline \multirow{4}{*}{8} & \multirow{4}{*}{48}
  & SEP & CUT & 5/5 & 1/5 & 1/5 & 1.222 &     - &  8562.8 &  926.6 & 500332.29 \\
& & SEP & SIM & 1/5 & 0/5 & 0/5 & 2.398 & 0.266 & 15236.8 & 1681.6 &  45498.12 \\
& & SVM & CUT & 4/5 & 0/5 & 0/5 & 1.559 & 0.018 & 12084.6 & 1310.0 & 496901.39 \\
& & SVM & SIM & 0/5 & 0/5 & 0/5 & 2.686 & 0.460 & 14751.0 & 1627.6 &  32216.24 \\
\hline \multirow{4}{*}{10} & \multirow{4}{*}{30}
  & SEP & CUT & 5/5 & 5/5 & 5/5 &     - &     - &  7485.8 &  705.6 &  56763.78 \\
& & SEP & SIM & 5/5 & 5/5 & 5/5 & 0.141 &     - & 18828.0 & 1819.6 &  35137.28 \\
& & SVM & CUT & 5/5 & 5/5 & 5/5 & 0.084 &     - & 12394.6 & 1176.4 &  78473.23 \\
& & SVM & SIM & 5/5 & 5/5 & 5/5 & 0.028 &     - & 16584.6 & 1594.0 &  31345.47 \\
\hline \multirow{4}{*}{10} & \multirow{4}{*}{40}
  & SEP & CUT & 5/5 & 5/5 & 5/5 & 0.167 &     - &  6735.0 &  608.8 & 248439.12 \\
& & SEP & SIM & 3/5 & 1/5 & 1/5 & 1.217 & 0.063 & 19417.8 & 1836.0 &  40114.23 \\
& & SVM & CUT & 4/5 & 5/5 & 5/5 & 0.336 & 0.021 & 13189.4 & 1228.8 & 351687.45 \\
& & SVM & SIM & 2/5 & 2/5 & 2/5 & 1.239 & 0.105 & 19331.6 & 1832.6 &  49527.99 \\
\hline \multirow{4}{*}{10} & \multirow{4}{*}{50}
  & SEP & CUT & 5/5 & 5/5 & 5/5 & 0.191 &     - & 10211.4 &  957.0 & 403198.82 \\
& & SEP & SIM & 4/5 & 1/5 & 1/5 & 1.099 & 0.068 & 19000.8 & 1812.4 &  21177.55 \\
& & SVM & CUT & 4/5 & 5/5 & 5/5 & 0.292 & 0.034 & 16001.8 & 1520.8 &  406346.30 \\
& & SVM & SIM & 4/5 & 2/5 & 2/5 & 0.926 & 0.017 & 19234.0 & 1828.0 &  33218.65 \\
\hline \multirow{4}{*}{10} & \multirow{4}{*}{60}
  & SEP & CUT & 5/5 & 5/5 & 5/5 & 0.208 &     - &  8687.8 &  781.4 & 426723.65 \\
& & SEP & SIM & 1/5 & 2/5 & 2/5 & 0.943 & 0.152 & 18810.4 & 1752.0 &  22610.29 \\
& & SVM & CUT & 5/5 & 5/5 & 5/5 & 0.250 &     - & 15133.4 & 1388.4 & 451552.54 \\
& & SVM & SIM & 1/5 & 4/5 & 4/5 & 0.818 & 0.111 & 19070.8 & 1783.0 &  32878.08 \\
\hline
\multicolumn{12}{l}{At each row -- column 5 to last -- statistics averaged over the specific set of instances are reported.}
\label{tab:gap_1}
\end{tabular}
\end{table}

\begin{table}[tb!]
\caption{Results of the experiments with instances of the Generalized Assignment Problem -- part 2.}
\centering 
\scriptsize
\begin{tabular}{c|c|l|l|r|r|r|r|r|r}
$m$ & $n$ & Separ. & Sampl. & \makecell{\# Calls to\\Thres.} & \makecell{\# Iters to\\Thres.} & \makecell{Time to\\Thres. (s)} & \makecell{\# Calls to\\Opt.} & \makecell{\# Iters to\\Opt.} & \makecell{Time to\\Opt. (s)} \\ 
\hline \multirow{4}{*}{5} & \multirow{4}{*}{15} 
  & SEP & CUT & 1202.6 & 227.4 & 471.44 & 299.8 & 39.0 & 75.73 \\
& & SEP & SIM & 1698.0 & 311.4 & 320.20 & 715.8 & 106.0 & 32.05 \\
& & SVM & CUT & 1607.2 & 290.4 & 741.50 & 578.8 & 80.2 & 222.24 \\
& & SVM & SIM & 1752.2 & 330.4 & 415.98 & 582.4 & 84.2 & 21.82 \\
\hline \multirow{4}{*}{5} & \multirow{4}{*}{20}
  & SEP & CUT & 999.4 & 158.0 & 1996.92 & 301.2 & 36.8 & 214.17 \\
& & SEP & SIM & 6401.8 & 1172.4 & 8065.87 & 1405.0 & 226.8 & 201.68 \\
& & SVM & CUT & 1503.0 & 241.0 & 2248.67 & 1115.4 & 172.4 & 1298.5 \\
& & SVM & SIM & 6296.0 & 1155.8 & 8715.71 & 1383.4 & 226.0 & 428.41 \\
\hline \multirow{4}{*}{5} & \multirow{4}{*}{25}
  & SEP & CUT & 732.8 & 98.4 & 1024.26 & 556.0 & 70.0 & 675.26 \\
& & SEP & SIM & 6503.0 & 1149.7 & 5846.29 & 3388.8 & 577.0 & 2445.27 \\
& & SVM & CUT & 1916.6 & 303.8 & 4391.99 & 1459.6 & 226.0 & 3280.67 \\
& & SVM & SIM & 6172.0 & 1089.0 & 4599.10 & 3540.8 & 595.6 & 2547.29 \\
\hline \multirow{4}{*}{5} & \multirow{4}{*}{30}
  & SEP & CUT & 1266.2 & 181.2 & 9199.39 & 653.8 & 80.4 & 1002.27 \\
& & SEP & SIM & - & - & - & 2024.0 & 278.0 & 293.96 \\
& & SVM & CUT & 2346.6 & 358.4 & 19084.71 & 1792.2 & 265.6 & 15946.49 \\
& & SVM & SIM & - & - & - & 3064.0 & 490.5 & 803.62 \\
\hline \multirow{4}{*}{8} & \multirow{4}{*}{24}
  & SEP & CUT & 1730.0 & 182.8 & 4826.99 & 803.4 & 72.8 & 967.42 \\
& & SEP & SIM & 4921.8 & 561.6 & 2302.02 & 1432.4 & 143.6 & 97.84 \\
& & SVM & CUT & 2719.6 & 290.6 & 5527.94 & 2023.8 & 209.4 & 3140.70 \\
& & SVM & SIM & 5001.2 & 571.2 & 2716.23 & 1232.2 & 122.0 & 98.39 \\
\hline \multirow{4}{*}{8} & \multirow{4}{*}{32}
  & SEP & CUT & 1468.0 & 143.0 & 3796.55 & 881.8 & 78.2 & 1182.11 \\
& & SEP & SIM & 10541.0 & 1239.0 & 7847.60 & 4124.8 & 448.2 & 1444.43 \\
& & SVM & CUT & 5184.4 & 574.0 & 19468.72 & 4938.6 & 551.2 & 48601.69 \\
& & SVM & SIM & - & - & - & 3086.3 & 325.5 & 841.83 \\
\hline \multirow{4}{*}{8} & \multirow{4}{*}{40}
  & SEP & CUT & 1877.4 & 180.8 & 12178.89 & 1457.2 & 133.8 & 3481.74 \\
& & SEP & SIM & 11234.5 & 1274.0 & 7555.32 & 7720.0 & 865.0 & 4599.56 \\
& & SVM & CUT & 5452.6 & 588.4 & 100765.94 & 6053.4 & 661.8 & 109843.42 \\
& & SVM & SIM & 4301.5 & 460.0 & 1047.74 & 7980.5 & 903.25 & 9003.12 \\
\hline \multirow{4}{*}{8} & \multirow{4}{*}{48}
  & SEP & CUT & 6228.0 & 649.0 & 327408.11 & 2291.8 & 194.0 & 14832.51 \\
& & SEP & SIM & - & - & - & 8700.0 & 911.0 & 6003.97 \\
& & SVM & CUT & - & - & - & 4267.0 & 388.5 & 26696.57 \\
& & SVM & SIM & - & - & - & - & - & - \\
\hline \multirow{4}{*}{10} & \multirow{4}{*}{30}
  & SEP & CUT & 1809.4 & 148.6 & 3029.43 & 1258.2 & 97.4 & 1644.78 \\
& & SEP & SIM & 8236.6 & 771.6 & 6545.39 & 4264.6 & 382.2 & 1484.13 \\
& & SVM & CUT & 5774.4 & 526.0 & 16064.81 & 6964.6 & 640.6 & 23693.84 \\
& & SVM & SIM & 7506.8 & 697.6 & 6255.76 & 3832.4 & 339.4 & 1908.10 \\
\hline \multirow{4}{*}{10} & \multirow{4}{*}{40}
  & SEP & CUT & 2316.6 & 182.2 & 13122.18 & 1779.0 & 133.2 & 4951.28 \\
& & SEP & SIM & 6920.0 & 625.0 & 1879.70 & 9717.33 & 890.0 & 9365.54 \\
& & SVM & CUT & 6918.0 & 618.2 & 51895.52 & 5472.0 & 481.5 & 23625.03 \\
& & SVM & SIM & 11060.5 & 1029.0 & 15175.82 & 7733.0 & 695.5 & 4341.77 \\
\hline \multirow{4}{*}{10} & \multirow{4}{*}{50}
  & SEP & CUT & 1917.0 & 152.6 & 9219.42 & 1806.8 & 141.2 & 4452.57 \\
& & SEP & SIM & 193.0 & 11.0 & 1.82 & 10954.0 & 1030.0 & 7281.48 \\
& & SVM & CUT & 4654.4 & 409.2 & 15939.45 & 5348.5 & 477.0 & 83367.05 \\
& & SVM & SIM & 5532.0 & 491.0 & 4121.31 & 10881.5 & 1017.0 & 7991.45 \\
\hline \multirow{4}{*}{10} & \multirow{4}{*}{60}
  & SEP & CUT & 1619.6 & 114.6 & 3243.72 & 2833.8 & 224.0 & 8940.38 \\
& & SEP & SIM & 1674.0 & 119.5 & 98.56 & 9077.0 & 798.0 & 4337.36 \\
& & SVM & CUT & 3349.8 & 262.6 & 8634.79 & 5053.6 & 419.0 & 17177.01 \\
& & SVM & SIM & 4725.0 & 401.8 & 2770.50 & 9452.0 & 853.0 & 4721.99 \\
\hline
\multicolumn{10}{l}{At each row -- column 5 to last -- statistics averaged over the specific set of instances are reported.}
\label{tab:gap_2}
\end{tabular}
\end{table}

We observe that SEP+CUT always yields the smallest Error, i.e., the best ending solution, for each class of problems, while SEP+SIM and SVM+SIM have the worst Error. In terms of Gap, SEP+CUT also has some advantage over the other approaches. These results demonstrate the strength of the proposed sampling strategy and the linear separation method, especially when they are combined. We also note that there is a significant difference between Gap and Error, especially for the Knapsack instances and the CSPP instances. This also shows that the quality of the lower bound (i.e., the obtained solutions) is much better than the quality of the conservative upper bound from Algorithm \ref{alg:ActiveLearnAndOptimize}. In terms of computing time, we observe that CUT can take much (sometimes orders of magnitude) longer time than SIM as it requires longer time to solve the MIQP \eqref{model:CUT}. If one has a budget on both the number of queries and computing time, then they have to carefully consider both when deciding which algorithm to use.

\section{Conclusions}
\label{sec:Conclusions}
In this paper, we proposed an interactive sampling-enhanced optimization framework and its alternative realizations, for solving combinatorial optimization problems with unknown knapsack constraints accessed by individual membership oracles. While such problems can be very hard in the worst case (as shown in Theorem \ref{thm:HardnessResult}), we have empirically shown that our framework can quickly find good primal solutions, especially when using the new sampling strategy with the linear separation method that we proposed. An interesting open question is if there exists any algorithm that matches the lower bound result of Theorem \ref{thm:HardnessResult}, or if this lower bound can be improved. On the other hand, as shown in the numerical experiments, the dual bound obtained from existing approaches seems to be much worse than the primal bound. This is maybe worth some further investigation.


\bibliographystyle{splncs04}
\bibliography{references}

\begin{thebibliography}{10}
\providecommand{\url}[1]{\texttt{#1}}
\providecommand{\urlprefix}{URL }
\providecommand{\doi}[1]{https://doi.org/#1}

\bibitem{aslan2023precedence}
Aslan, A., Ursavas, E., Romeijnders, W.: A precedence constrained knapsack problem with uncertain item weights for personalized learning systems. Omega  \textbf{115},  102779 (2023)

\bibitem{awasthi2015efficient}
Awasthi, P., Balcan, M.F., Haghtalab, N., Urner, R.: Efficient learning of linear separators under bounded noise. In: Conference on Learning Theory. pp. 167--190. PMLR (2015)

\bibitem{beasley1990or}
Beasley, J.E.: Or-library, generalized assignment problem (\url{http://people.brunel.ac.uk/~mastjjb/jeb/orlib/gapinfo.html}) (1990)

\bibitem{bellman1957comment}
Bellman, R.: Comment on dantzig's paper on discrete variable extremum problems. Operations Research  \textbf{5}(5),  723--724 (1957)

\bibitem{ben2009robust}
Ben-Tal, A., Nemirovski, A., El~Ghaoui, L.: Robust optimization  (2009)

\bibitem{bertsimas2004solving}
Bertsimas, D., Vempala, S.: Solving convex programs by random walks. Journal of the ACM (JACM)  \textbf{51}(4),  540--556 (2004)

\bibitem{bessiere2023learning}
Bessiere, C., Carbonnel, C., Dries, A., Hebrard, E., Katsirelos, G., Narodytska, N., Quimper, C.G., Stergiou, K., Tsouros, D.C., Walsh, T.: Learning constraints through partial queries. Artificial Intelligence  \textbf{319},  103896 (2023)

\bibitem{birnbaum2012learning}
Birnbaum, A., Shwartz, S.: Learning halfspaces with the zero-one loss: time-accuracy tradeoffs. Advances in Neural Information Processing Systems  \textbf{25} (2012)

\bibitem{blum1998polynomial}
Blum, A., Frieze, A., Kannan, R., Vempala, S.: A polynomial-time algorithm for learning noisy linear threshold functions. Algorithmica  \textbf{22},  35--52 (1998)

\bibitem{boser1992training}
Boser, B.E., Guyon, I.M., Vapnik, V.N.: A training algorithm for optimal margin classifiers. In: Proceedings of the fifth annual workshop on Computational learning theory. pp. 144--152 (1992)

\bibitem{cacchiani2022knapsackA}
Cacchiani, V., Iori, M., Locatelli, A., Martello, S.: Knapsack problems — an overview of recent advances. part i: single knapsack problems. Computers \& Operations Research  \textbf{143},  105692 (2022)

\bibitem{cacchiani2022knapsackB}
Cacchiani, V., Iori, M., Locatelli, A., Martello, S.: Knapsack problems — an overview of recent advances. part ii: Multiple, multidimensional, and quadratic knapsack problems. Computers \& Operations Research  \textbf{143},  105693 (2022)

\bibitem{ECSurvey}
Cao, L., Huo, T., Li, S., Zhang, X., Chen, Y., Lin, G., Wu, F., Ling, Y., Zhou, Y., Xie, Q.: Cost optimization in edge computing: a survey. Artificial Intelligence Review  \textbf{57} (2024)

\bibitem{cattrysse1992survey}
Cattrysse, D.G., Van~Wassenhove, L.N.: A survey of algorithms for the generalized assignment problem. European journal of operational research  \textbf{60}(3),  260--272 (1992)

\bibitem{chan2025inverse}
Chan, T.C., Mahmood, R., Zhu, I.Y.: Inverse optimization: Theory and applications. Operations Research  \textbf{73}(2),  1046--1074 (2025)

\bibitem{cheney1959newton}
Cheney, E.W., Goldstein, A.A.: Newton's method for convex programming and tchebycheff approximation. Numerische Mathematik  \textbf{1}(1),  253--268 (1959)

\bibitem{dadush2023simple}
Dadush, D., Hojny, C., Huiberts, S., Weltge, S.: A simple method for convex optimization in the oracle model. Mathematical Programming pp. 1--22 (2023)

\bibitem{daniely2015ptas}
Daniely, A.: A ptas for agnostically learning halfspaces. In: Conference on Learning Theory. pp. 484--502. PMLR (2015)

\bibitem{daniely2016complexity}
Daniely, A.: Complexity theoretic limitations on learning halfspaces. In: Proceedings of the forty-eighth annual ACM symposium on Theory of Computing. pp. 105--117 (2016)

\bibitem{dantzig1957discrete}
Dantzig, G.B.: Discrete-variable extremum problems. Operations research  \textbf{5}(2),  266--288 (1957)

\bibitem{fajemisin2024optimization}
Fajemisin, A.O., Maragno, D., den Hertog, D.: Optimization with constraint learning: A framework and survey. European Journal of Operational Research  \textbf{314}(1),  1--14 (2024)

\bibitem{fisher1986multiplier}
Fisher, M.L., Jaikumar, R., Van~Wassenhove, L.N.: A multiplier adjustment method for the generalized assignment problem. Management science  \textbf{32}(9),  1095--1103 (1986)

\bibitem{fouskakis2002stochastic}
Fouskakis, D., Draper, D.: Stochastic optimization: a review. International Statistical Review  \textbf{70}(3),  315--349 (2002)

\bibitem{fu2013survey}
Fu, Y., Zhu, X., Li, B.: A survey on instance selection for active learning. Knowledge and information systems  \textbf{35},  249--283 (2013)

\bibitem{goffin1993computation}
Goffin, J.L., Vial, J.P.: On the computation of weighted analytic centers and dual ellipsoids with the projective algorithm. Mathematical Programming  \textbf{60}(1),  81--92 (1993)

\bibitem{gomory2010outline}
Gomory, R.E.: An algorithm for integer solutions to linear programs. Bulletin of the American Mathematical Society  \textbf{64},  275–--278 (1958)

\bibitem{grotschel2012geometric}
Gr{\"o}tschel, M., Lov{\'a}sz, L., Schrijver, A.: Geometric algorithms and combinatorial optimization, vol.~2. Springer Science \& Business Media (2012)

\bibitem{gurobi}
{Gurobi Optimization, LLC}: {Gurobi Optimizer Reference Manual} (2023), \url{https://www.gurobi.com}

\bibitem{guruswami2009hardness}
Guruswami, V., Raghavendra, P.: Hardness of learning halfspaces with noise. SIAM Journal on Computing  \textbf{39}(2),  742--765 (2009)

\bibitem{hino2020active}
Hino, H.: Active learning: Problem settings and recent developments. arXiv preprint arXiv:2012.04225  (2020)

\bibitem{ho2011active}
Ho, C.H., Tsai, M.H., Lin, C.J.: Active learning and experimental design with svms. In: Active Learning and Experimental Design workshop In conjunction with AISTATS 2010. pp. 71--84. JMLR Workshop and Conference Proceedings (2011)

\bibitem{jiang2020improved}
Jiang, H., Lee, Y.T., Song, Z., Wong, S.C.w.: An improved cutting plane method for convex optimization, convex-concave games, and its applications. In: Proceedings of the 52nd Annual ACM SIGACT Symposium on Theory of Computing. pp. 944--953 (2020)

\bibitem{kelley1960cutting}
Kelley, Jr, J.E.: The cutting-plane method for solving convex programs. Journal of the Society for Industrial and Applied Mathematics  \textbf{8}(4),  703--712 (1960)

\bibitem{kremer2014active}
Kremer, J., Steenstrup~Pedersen, K., Igel, C.: Active learning with support vector machines. Wiley Interdisciplinary Reviews: Data Mining and Knowledge Discovery  \textbf{4}(4),  313--326 (2014)

\bibitem{kudla2018one}
Kud{\l}a, P., Pawlak, T.P.: One-class synthesis of constraints for mixed-integer linear programming with c4. 5 decision trees. Applied Soft Computing  \textbf{68},  1--12 (2018)

\bibitem{kumar2019acquiring}
Kumar, M.: Acquiring integer programs from data. In: Proceedings of the Twenty-Eighth International Joint Conference on Artificial Intelligence. pp. 1130--1136. Ijcai (2019)

\bibitem{kumar2020active}
Kumar, P., Gupta, A.: Active learning query strategies for classification, regression, and clustering: a survey. Journal of Computer Science and Technology  \textbf{35},  913--945 (2020)

\bibitem{lee2018efficient}
Lee, Y.T., Sidford, A., Vempala, S.S.: Efficient convex optimization with membership oracles. In: Conference On Learning Theory. pp. 1292--1294. PMLR (2018)

\bibitem{lee2015faster}
Lee, Y.T., Sidford, A., Wong, S.C.w.: A faster cutting plane method and its implications for combinatorial and convex optimization. In: 2015 IEEE 56th Annual Symposium on Foundations of Computer Science. pp. 1049--1065. IEEE (2015)

\bibitem{lombardi2018boosting}
Lombardi, M., Milano, M.: Boosting combinatorial problem modeling with machine learning. Proceedings of the Twenty-Seventh International Joint Conference on Artificial Intelligence  (2018)

\bibitem{mammone2009support}
Mammone, A., Turchi, M., Cristianini, N.: Support vector machines. Wiley Interdisciplinary Reviews: Computational Statistics  \textbf{1}(3),  283--289 (2009)

\bibitem{martello1990knapsack}
Martello, S., Toth, P.: Knapsack problems: algorithms and computer implementations. John Wiley \& Sons, Inc. (1990)

\bibitem{mathews1896partition}
Mathews, G.B.: On the partition of numbers. Proceedings of the London Mathematical Society  \textbf{1}(1),  486--490 (1896)

\bibitem{mechqrane2024using}
Mechqrane, Y., Bessiere, C., Elabbassi, I.: Using large language models to improve query-based constraint acquisition. In: IJCAI 2024-33rd International Joint Conference on Artificial Intelligence. pp. 1916--1925 (2024)

\bibitem{mitchell1982generalization}
Mitchell, T.M.: Generalization as search. Artificial intelligence  \textbf{18}(2),  203--226 (1982)

\bibitem{nesterov1995cutting}
Nesterov, Y.: Cutting plane algorithms from analytic centers: efficiency estimates. Mathematical Programming  \textbf{69}(1),  149--176 (1995)

\bibitem{oncan2007survey}
{\"O}ncan, T.: A survey of the generalized assignment problem and its applications. INFOR: Information Systems and Operational Research  \textbf{45}(3),  123--141 (2007)

\bibitem{pawlak2017automatic}
Pawlak, T.P., Krawiec, K.: Automatic synthesis of constraints from examples using mixed integer linear programming. European Journal of Operational Research  \textbf{261}(3),  1141--1157 (2017)

\bibitem{pisinger1999core}
Pisinger, D.: Core problems in knapsack algorithms. Operations Research  \textbf{47}(4),  570--575 (1999)

\bibitem{powell2019unified}
Powell, W.B.: A unified framework for stochastic optimization. European journal of operational research  \textbf{275}(3),  795--821 (2019)

\bibitem{QUADRI202330}
Quadri, C., Ceselli, A., Rossi, G.P.: Multi-user edge service orchestration based on deep reinforcement learning. Computer Communications  \textbf{203},  30--47 (2023). \doi{https://doi.org/10.1016/j.comcom.2023.02.027}, \url{https://www.sciencedirect.com/science/article/pii/S0140366423000737}

\bibitem{rosenblatt1958perceptron}
Rosenblatt, F.: The perceptron: a probabilistic model for information storage and organization in the brain. Psychological review  \textbf{65}(6), ~386 (1958)

\bibitem{ross1975branch}
Ross, G.T., Soland, R.M.: A branch and bound algorithm for the generalized assignment problem. Mathematical programming  \textbf{8}(1),  91--103 (1975)

\bibitem{schohn2000less}
Schohn, G., Cohn, D.: Less is more: Active learning with support vector machines. In: Proceedings of the Seventeenth International Conference on Machine Learning. p. 839–846. ICML '00, Morgan Kaufmann Publishers Inc., San Francisco, CA, USA (2000)

\bibitem{settles2009active}
Settles, B.: Active learning literature survey. Computer Sciences Technical Report~1648, University of Wisconsin--Madison (2009)

\bibitem{shalev2011learning}
Shalev-Shwartz, S., Shamir, O., Sridharan, K.: Learning kernel-based halfspaces with the 0-1 loss. SIAM Journal on Computing  \textbf{40}(6),  1623--1646 (2011)

\bibitem{tharwat2023survey}
Tharwat, A., Schenck, W.: A survey on active learning: state-of-the-art, practical challenges and research directions. Mathematics  \textbf{11}(4), ~820 (2023)

\bibitem{tong2001active}
Tong, S.: Active learning: theory and applications. Stanford University (2001)

\bibitem{tong2001support}
Tong, S., Koller, D.: Support vector machine active learning with applications to text classification. Journal of machine learning research  \textbf{2}(Nov),  45--66 (2001)

\bibitem{vaidya1996new}
Vaidya, P.M.: A new algorithm for minimizing convex functions over convex sets. Mathematical programming  \textbf{73}(3),  291--341 (1996)

\bibitem{vapnik1995support}
Vapnik, V.: Support-vector networks. Machine learning  \textbf{20},  273--297 (1995)

\bibitem{xu2003representative}
Xu, Z., Yu, K., Tresp, V., Xu, X., Wang, J.: Representative sampling for text classification using support vector machines. In: Advances in Information Retrieval: 25th European Conference on IR Research, 2003. Proceedings 25. pp. 393--407. Springer (2003)

\bibitem{zheng2017active}
Zheng, S., Waggoner, B., Liu, Y., Chen, Y.: Active information acquisition for linear optimization. Conference on Uncertainty in Artificial Intelligence  (2018)

\end{thebibliography}

\appendix

\section{Proofs}
\label{sec:Proofs}

Theorem \ref{thm:HardnessResult} shows that finding an optimal solution to Problem \eqref{model:01LinearProgramWithKnapsackConstraints} with a single unknown constraint can require enumerating a number of solutions exponential in the number of variables $n$.

\begin{proof}[Theorem \ref{thm:HardnessResult}]
    For any fixed $\epsilon\in(0,1)$, consider the instance $\mathcal{I}$ of the problem defined by
    \begin{center}
        $\boldsymbol{v} = \boldsymbol{1}$,\ \ \ $\bar{\boldsymbol{w}} = \dfrac{\boldsymbol{1}}{C}$,
    \end{center}
    where $C = \lceil 1 / \epsilon\rceil - 1$. The feasible solutions are those with at most $C$ items equal to 1. In particular, all the solutions picking $C$ items equal to 1 are feasible and all those picking $C + 1$ items equal to 1 are infeasible.
    Consider now the set $\mathcal{P}$ composed of instances of the problem that satisfy
    \begin{center}
        $\boldsymbol{v}^{\prime} = \boldsymbol{1}$,\ \ \ $\bar{w}_j^{\prime} = 
        \begin{cases}
            \dfrac{1}{C+1} & \text{for a subset of } C+1 \text{ items}, \\
            \dfrac{1}{C} & \text{for all other items.}
        \end{cases}$
    \end{center}
    Each instance in $\mathcal{P}$ has a unique optimal solution by picking all the items with weight $1 / (C+1)$, which attains the optimal objective value $\lceil 1 / \epsilon\rceil=C+1$. All other solutions picking at least $C+1$ items equal to 1 are infeasible and all the solutions picking at most $C$ items equal to 1 are feasible. For every $\mathcal{I}^{\prime} \in \mathcal{P}$, the feasible sets for instance $\mathcal{I}$ and $\mathcal{I}^{\prime}$ coincide except that the set for $\mathcal{I}^{\prime}$ contains an additional solution, which is also optimal.

    Note that, for a deterministic algorithm, the solution to be labeled can only be determined by the previously labeled solutions.
    Since the solution algorithm is not aware of the actual item weights, it cannot distinguish between $\mathcal{I}$ and any instance $\mathcal{I}^{\prime} \in \mathcal{P}$ unless the optimal solution of $\mathcal{I}^{\prime}$ is labeled. In particular, this is true even if the algorithm has already labeled an optimal solution for $\mathcal{I}$. For instance $\mathcal{I}$, we assume the solution algorithm does not terminate before it labels all solutions picking $C+1$ items (since before that the algorithm cannot guarantee the optimality of any labeled solution). Then, given the sequence of solutions that the algorithm labels for $\mathcal{I}$, one can construct an instance $\mathcal{I}^{\prime\prime} \in \mathcal{P}$ that runs in exponential time for the algorithm. Let $\tilde{x}$ be the lastly labeled solution that picks $C+1$ items. Define the instance $\mathcal{I}^{\prime\prime}\in \mathcal{P}$ as follows:
    \begin{center}
        $\boldsymbol{v}^{\prime\prime} = \boldsymbol{1}$,\ \ \ $\bar{w}_j^{\prime\prime} = 
        \begin{cases}
            \dfrac{1}{C + 1} & \text{if } \tilde{x}_j = 1, \\
            \dfrac{1}{C} & \text{if } \tilde{x}_j = 0.
        \end{cases}$
    \end{center}
    By definition, $\tilde{x}$ is the unique better-than-(1-$\epsilon$)-optimal solution of $\mathcal{I}^{\prime\prime}$, and is the last labeled solution among all the solutions that pick $C+1$ items. Labeling it can require the algorithm to at least label all the solutions picking $C+1=\lceil 1/\epsilon \rceil$ items. In this case, the algorithm requires at least ${{n}\choose {\lceil 1/\epsilon \rceil}}=\Omega\left(n^{1/\epsilon}\right)$ calls to the membership oracle. \qed
\end{proof}

Proposition \ref{prop:Bounding} shows that any optimal solution of Problem \eqref{model:Bounding} has a value that is greater than or equal to that of any optimal solution of Problem \eqref{model:01LinearProgramWithKnapsackConstraints}.

\begin{proof}[Proposition \ref{prop:Bounding}]
For every optimal solution $\boldsymbol{x}^{\circledast}$ of Problem \eqref{model:01LinearProgramWithKnapsackConstraints}, a $\boldsymbol{w}^{\prime} \in W$ exists such that $(\boldsymbol{x}^{\circledast}, \boldsymbol{w}^{\prime})$ is feasible for Problem \eqref{model:Bounding} - possibly not optimal. In fact, let us take $\boldsymbol{w}^{\prime} = \bar{\boldsymbol{w}}$, the parameters of constraints \eqref{constr:KnapsackConstraints}:
\begin{enumerate}
    \item [$\cdot$] for each $i \in I$, the $i$-th constraint in \eqref{constr:BoundingKnapsackConstraints} is satisfied because the corresponding constraint in \eqref{constr:KnapsackConstraints} is;
    \item [$\cdot$] for each $i \in I$ and for each $\boldsymbol{\sigma}^+ \in S^+_i(t)$, the corresponding constraint in \eqref{constr:BoundingFeasiblePoints} becomes
    \begin{equation}
        \sum_{j=1}^n \bar{w}_{ij} \cdot \sigma^+_j \leq 1\text{.}
    \end{equation}
    This constraint corresponds to the $i$-th constraint in \eqref{constr:KnapsackConstraints} evaluated in $\boldsymbol{x}_i = \boldsymbol{\sigma}^+$, which results in a valid inequality because $\boldsymbol{\sigma}^+$ is labeled as a positive point;
    \item [$\cdot$] similarly, for each $i \in I$ and for each $\boldsymbol{\sigma}^- \in S^-_i(t)$, the corresponding constraint in \eqref{constr:BoundingInfeasiblePoints} is satisfied because $\boldsymbol{\sigma}^-$ is labeled as an infeasible point and, therefore, it violates the $i$-th constraint in \eqref{constr:KnapsackConstraints};
    \item [$\cdot$] constraints \eqref{constr:BoundingNoGoodCuts} state that for each $i \in I$ and for each $\boldsymbol{\sigma}^- \in S^-_i(t)$, the sub-point $\boldsymbol{x}_i$ of $\boldsymbol{x}$ is different from $\boldsymbol{\sigma}^-$. This is true for $\boldsymbol{x} = \boldsymbol{x}^{\circledast}$ because $\boldsymbol{\sigma}^-$ is labeled as negative with respect to the $i$-th constraint in \eqref{constr:KnapsackConstraints} and, therefore, if $\boldsymbol{x}^{\circledast}_i$ was equal to $\boldsymbol{\sigma}^-$, it would violate the $i$-th constraint in \eqref{constr:KnapsackConstraints};
    \item [$\cdot$] constraint \eqref{constr:BoundingCombinatorial} is satisfied because it is the same as constraint \eqref{constr:Combinatorial};
    \item [$\cdot$] finally, constraint \eqref{constr:BoundingParameterDomain} is satisfied because $\bar{\boldsymbol{w}} \in W$ by definition.
\end{enumerate}
We can conclude that, for every optimal solution $(\boldsymbol{x}^*, \boldsymbol{w}^*)$ of Problem \eqref{model:Bounding}, it holds that, according to its optimality, $z(\boldsymbol{x}^*) \geq z(\boldsymbol{x}^{\circledast})$. \qed
\end{proof}

\section{Application Example}
\label{apd:application}
We report an application scenario fitting the setting of the paper. It is related to optimal orchestration problems in edge computing. Edge computing networks are designed for supporting end users that have devices with limited hardware, but need to perform tasks requiring high computing effort and low latency. A common example is augmented reality: end users have mobile devices whose CPU, RAM and, most of all, battery are limited; they ask for image processing capabilities that can be obtained only by \emph{offloading} tasks, that is, running them in virtual machines on remote servers instead of locally on the device. Augmented reality tasks need to be highly responsive: offloading to virtual machines in the cloud would produce lag and overall a poor quality of service. The philosophy of edge computing is to exploit a set of smaller servers, installed at the `edge' of the network (e.g., where base stations antennas are placed). Clearly, their computing resources can be much larger than those of mobile devices, but not as large as far cloud facilities: a careful planning needs to be done. A network service named \emph{orchestrator} is in charge of optimally managing their resources. Among its duties, an orchestrator needs to choose to which server to allocate the tasks of the users. A greedy choice would be to allocate each task to the server providing highest quality of service (e.g., lowest latency). However, by doing so, some servers might become overloaded, and therefore an optimal balancing needs to be performed \cite{ECSurvey}.

A fundamental issue is the following: the resource consumption of user tasks are unknown to the orchestrator \cite{QUADRI202330}. Therefore the orchestrator might perform an allocation decision, which is detected to be infeasible after its implementation. For instance, the orchestrator might allocate a set of tasks to the same server, detecting after allocation that they are consuming more RAM than that available on the server. The orchestrator is allowed to change its allocation decisions. However, each reallocation requires the migration of virtual machines from one server to another, introducing delays and load on the network.

The setting of our paper helps in the automatic decisions of these orchestrators. Let $J$ be a set of tasks whose computation is required. Let $I$ be a set of servers. Let $v_{ij}$ a measure of quality of service for allocating task $j \in J$ to server $i \in I$. Let $\bar w_{ij}$ be a relative resource consumption of task $j \in J$ while running on server $i \in I$. The aim of the orchestrator is to distribute tasks to the servers, optimizing the overall quality of service, in such a way that the relative resource consumption on each server does not exceed the server capacity. It can be modeled as ~\eqref{obj:01LinearProgramWithKnapsackConstraints} -- \eqref{constr:Combinatorial}. 

The relative resource consumption of each task is initially unknown. The orchestrator might operate as follows. It allocates tasks to servers in a greedy way. Some servers will possibly be overloaded, and start violating some feasibility metric (e.g., number of page faults, number of lost packets, avg. core temperature). This is seen as an `infeasibility' found by a membership oracle in the corresponding constraint. Some tentative feasible allocations may also be obtained by various policies and heuristics, possibly at the expense of poor overall quality of service to the users. The orchestrator is allowed to reallocate tasks, exploiting the hyperplane produced by the oracle to populate a mathematical program, and optimizing it instead of performing a greedy allocation. Such a reallocation and feasibility check might be performed with a limited frequency (e.g., at most a given number of times per batch of a single minute).

\end{document}